\pdfoutput=1

\documentclass[11pt]{article}
\usepackage{float}
\usepackage[table]{xcolor} 
\usepackage{booktabs} 
\usepackage{adjustbox}
\usepackage{tabularx}
\usepackage{subcaption}
\usepackage{array}
\usepackage{xcolor}
\usepackage{caption}
\usepackage{hyperref}
\usepackage{ragged2e}
\newcolumntype{Y}{>{\RaggedRight\arraybackslash}X}
\usepackage{adjustbox} 
\usepackage{appendix}

\usepackage[preprint]{acl}

\usepackage{times}
\usepackage{latexsym}

\usepackage[T1]{fontenc}

\usepackage[utf8]{inputenc}

\usepackage{microtype}

\usepackage{inconsolata}

\usepackage{graphicx}
\usepackage{array}
\usepackage{amsmath} 

\title{RideKE: Leveraging Low-Resource, User-Generated Twitter Content for Sentiment and Emotion Detection in Kenya\`{n} Code-Switched Dataset}

\author{Naome A. Etori\ and Maria L. Gini \\ 
  Department of Computer Science and Engineering  \\
  University of Minnesota -Twin Cities\\
  \texttt\{etori001, gini\} @umn.edu}

\begin{document}
\maketitle
\begin{abstract}
Social media has become a crucial open-access platform for individuals to express opinions and share experiences. 
However, leveraging low-resource language data from Twitter is challenging due to scarce, poor-quality content and the major variations in language use, such as  slang and code-switching.  Identifying tweets in these languages can be difficult as Twitter primarily supports high-resource languages. 
We analyze Kenyan code-switched data and evaluate four state-of-the-art (SOTA) transformer-based pretrained models for sentiment and emotion classification, using supervised and semi-supervised methods. We detail the methodology behind data collection and annotation, and the challenges encountered during the data curation phase. Our results show that XLM-R outperforms other models; for sentiment analysis, XLM-R supervised model achieves the highest accuracy (69.2\%) and F1 score (66.1\%), XLM-R semi-supervised (67.2\% accuracy, 64.1\% F1 score). In emotion analysis, DistilBERT supervised leads in accuracy (59.8\%) and F1 score (31\%), mBERT semi-supervised (accuracy (59\% and F1 score 26.5\%). AfriBERTa models show the lowest accuracy and F1 scores. All models tend to predict neutral sentiment, with Afri-BERT showing the highest bias and unique sensitivity to empathy emotion. \footnote{\url{https://github.com/NEtori21/Ride_hailing_project}}

\end{abstract}

\section{Introduction}

\begin{figure}[ht]
    \centering
    \includegraphics[width=\linewidth]{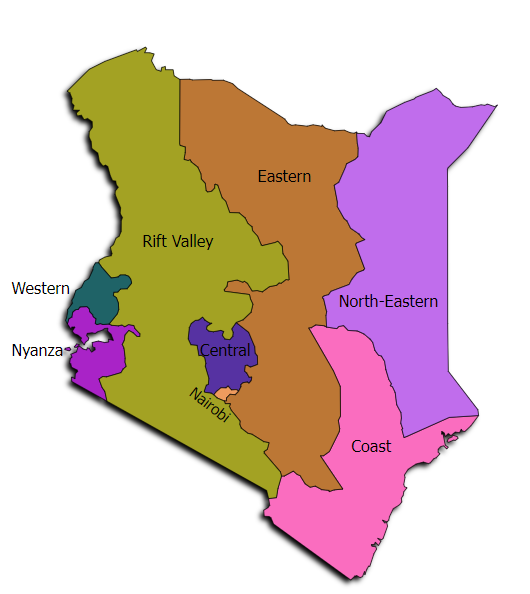} 
    \caption {\textbf{Geographical representation of RideKE:} 
    \textit{diverse local accents collected in tweets, such as Rift Valley (e.g., Eldoret, Nakuru), Central (e.g., Nyeri, Kiambu), Nairobi (e.g., Kasarani, Kileleshwa), Western (e.g., Kakamega, Bungoma), Nyanza (e.g., Kisumu, Kisii), Eastern (e.g., Machakos, Embu) Coast (e.g., Mombasa, Malindi), and North-Eastern (e.g., Garissa, Mandera).}}
    \label{fig:ken}
\end{figure}

\begin{table*}[t!]
  \small
  \centering
  \begin{tabular}{p{1.4\columnwidth}cc}
    \hline
    \textbf{Tweets} & \textbf{Sentiment} & \textbf{Emotion} \\
    \hline
    \rowcolor{gray!25}
    Uber kenya did your App stop accepting cards for package deliveries? I have had two riders this morning cancel picking a package because they want me to pay cash. & Negative & Frustration \\
    Thank you for the love and support and for the feedback as well. Tell all your friends to ride a littleCab. Buy Kenyan, build Kenya. & Positive & Love \\
    \rowcolor{gray!25}
    Uber drivers are not employees of Uber Kenya Uber is only an app. The link between you as a rider and the driver. But yes they should look after them because the drivers keep them afloat. & Neutral & Neutral \\
    A ride will be canceled for one reason or another and both parties should have the liberty to. Sometimes clients will cancel due to the proximity of the driver and other times because the driver is unreachable. & Neutral & Neutral \\
    \rowcolor{gray!25}
    Hope everyone making the most of this awesome Uber kenya Jan offer! Spread the word! Loving it. \#Uber kenya & Positive & Happy \\
    Giving drivers right to refer the rider to another driver then that is totally not a good idea. Some drivers are connecting while he like really far from you, he wastes time, then after more than 5 mins refers another driver & Neutral & Happy \\
    \rowcolor{gray!25}
    Greater experience for Uber riders with new product & Positive & Happy \\
    I am reporting your driver for taking payment twice. I had ordered an Uber for a friend with payment with a card and then he tells the passenger to pay via Mpesa. & Negative & Frustration \\
    \rowcolor{gray!25}
    I also stopped using Uber kenya after I was charged for cancelling a trip as per the drivers request. Little cab iko tu sawa. & Negative & Frustration \\
    Crooked policies. Uber kenya. I think you need to sort out your service. & Negative & Angry \\
    \rowcolor{gray!25}
    Honestly, Am disappointed with them. kucancel trips ndio wanajua lately. & Negative & Frustration \\
    \hline
  \end{tabular}
  \caption{Sample Tweets with Sentiment and Emotion Labels.}
  \label{tab:tweets_sentiment_emotion}
\end{table*}

Kenya, reflecting Africa's extensive multilingual diversity, offers a unique insight into the continent's rich linguistic heritage, standing as a focal point of language contact, expansion, and diversity. It is home to many languages that bridge its vibrant storytelling, poetry, song, and literature and exemplifies Africa's linguistic wealth, albeit on a more localized scale. With over 40 languages grouped into Bantu, Nilotic, and Cushitic, Kenya's linguistic landscape is diverse and dynamic \cite{dwivedi2014linguistic,carter2007teaching,banks2002talk}.

Central to linguistic diversity is the co-official language status of English and Kiswahili, with the latter spoken by the majority and enjoying near-equal prominence with English. However, the linguistic equilibrium faces challenges from Sheng, a language that blends English, Kiswahili, and words from other ethnic languages that initially were used in Nairobi Eastlands slums. Sheng emerged as a sociolect among urban youth in the city's working-class neighborhoods and has since spread across various social and age groups. Hence, it is an integral part of Kenyan culture, influencing the traditional dominance of English and Kiswahili \cite{barasa2016spoken, momanyi2009effects,mazrui1995slang}. 

In recent years, language diversity has also been mirrored in the urban transportation sector, primarily due to the growth of Ride-Hailing Services (RHS) such as Uber, Bolt, and Little Cab. These services have rapidly transformed from urban novelties to essential components of daily mobility for many Kenyans, connecting remote areas with vibrant urban cities. 
However, with the entry of global giants like Uber in 2015, followed by Bolt and the local contender Little Cab, this transformation is not just physical; it extends into digital and social media platforms such as Twitter.

Since many  languages are spoken across Kenya, each population has its own dialect. Hence, code-switching is common in these new forms of communication, where speakers alternate between two or more languages in one conversation \cite{kanana2018functions,santy2021bertologicomix,angel2020nlp,thara2018code}. Analyzing sentiment and emotions in code-switched language context is critical in the broad natural language processing  (NLP) field, for example, creating systems that can predict emotional states from text to speech which can be applied in various use cases, such as measuring consumer satisfaction \cite{ren2012linguistic}, natural disasters \cite{vo2013twitter}, marketing strategy \cite{zamani2016eye}, e-learning \cite{ortigosa2014sentiment}, e-commerce\cite{jabbar2019real} and psychological states \cite{aytuug2018sentiment}. However, despite this linguistic richness, African languages remain significantly underrepresented in NLP research \cite{muhammad2023afrisenti}. Although NLP research has made extensive progress and demonstrated broad utility over the past two decades, the focus on African languages has been limited. This disparity is often attributed to the scarcity of high-quality, annotated datasets for these languages.

Recently, researchers \cite{muhammad2023afrisenti}\footnote{\url{https://github.com/afrisenti-semeval/afrisent-semeval-2023}} have focused on addressing this challenge by introducing a comprehensive benchmark with over 110,000 tweets across 14 African languages, Swahili among them, and introduced the first Africentric SemEval Shared task \cite{muhammad2023semeval}. Various studies have evaluated the performance of state-of-the-art (SOTA) transformer models on African languages, highlighting unique challenges and opportunities \cite{aryal2023sentiment}.

\begin{table*}[ht!]
\centering
\footnotesize
\begin{tabularx}{\textwidth}{Y Y}
\toprule
\textbf{Code-switched Reference} & \textbf{English Translation} \\
\midrule
I recently interacted with one Uber driver who told me that \textbf{\textcolor{blue}{huko ni mbali, lazima uongeze pesa}}. Different from the estimate on the app. He almost dropped me midway because I argued that it wasn't fair. \textbf{\textcolor{blue}{Hawa madere ni wazimu walai.}} & I recently interacted with one Uber driver who told me that \underline{\textcolor{red}{the place is far, you have to add money}}. Different from the estimate on the app. He almost dropped me midway because I argued that it wasn't fair. \underline{\textcolor{red}{These drivers are crazy, really.}} \\
\midrule
In Mombasa, they ask you how much the App has displayed as the cost, then tell you it's too low, madam \textbf{\textcolor{blue}{unaona utaongeza ngapi}}, \textbf{\textcolor{blue}{hiyo pesa ni kidogo}} & In Mombasa they ask you how much the app has displayed as the cost then tell you it's too low, madam \underline{\textcolor{red}{how much extra?}}, \underline{\textcolor{red}{That's little money}} \\
\bottomrule
\end{tabularx}
\caption{Example of code-switched sentences in Tweets}
\label{tab:codeswitched}
\end{table*}

However, research on social media NLP analysis for RHS datasets mainly targets high-resource languages. NLP for low-resource languages is constrained by factors like NLP research's geographical and language diversity \cite{joshi2020state}. Using pre-trained transformer models, we introduce RideKE, a sentiment and emotion analysis dataset for  African-accented English code switched with Swahili and Sheng. 

Our dataset contains over 29,000 tweets, each sentiment classified as either positive, negative, or neutral, and emotions classified as frustration, happy, angry, sad, empathy, fear, love, and surprise. The dataset represents one location, Kenya, as shown in Table~\ref{tab:tweets_sentiment_emotion}. Our goal is to advance research in low-resource languages.

The experiments in this paper are designed to allow us to answer the following specific questions:

\begin{enumerate}
    \item How do pretrained language models enhance the detection and representation of Kenyan low-resource languages and accents in modern NLP tools?
    \item How does the performance of sentiment and emotion detection varies across different pretrained transformer-based models?
    \item How effective are different transformer-based models in performing sentiment and emotion detection on the low-resource (RideKE) dataset using semi-supervised learning?
\end{enumerate}

Our paper makes the following contributions as we address these questions:

\begin{itemize}\setlength{\itemsep}{1pt} \setlength{\parskip}{0pt \setlength{\parsep}{0pt}}
    \item  We use semi-supervised learning to classify sentiments and emotions. We compare four SOTA transformer-based models and provide a detailed model performance analysis.
    \item We contribute a partially curated human-annotated labeled public dataset with over 29,000 tweets from the RHS domain. This is Kenya's first-ever code-switched sentiment and emotion dataset in the RHS domain. It contributes resources to low-resource areas, which can be used for other analyses.
\end{itemize}

\section{Literature Review}
\label{sxn:rw}
\subsection{Sentiment Analysis on Social Media}
Sentiment analysis (SA) emerged as a significant field early in the 2000s \cite{das2001yahoo, nasukawa2003sentiment}.  
SA \cite{dave2003mining, pang2008opinion} aims to determine the attitudes, opinions, or emotions expressed in text on specific topics or entities \cite{liu2022sentiment} and has become an increasingly popular research area. Due to higher user-generated content available on social media, understanding  sentiment in text cannot be overstated \cite{naseem2019dice}. 

Diverse strategies to accurately interpret and classify user sentiments have been employed. For example, lexicon-based approaches, like SENTIWORDNET \cite{baccianella2010sentiwordnet} and AFINN \cite{nielsen2011new}, used predefined word lists to classify text sentiment. While effective in some applications, these methods often struggled with context and nuance. Rule-based systems \cite{suttles2013distant} further enhanced this method by applying contextual rules to detect sentiment nuances, including handling negations \cite{taboada2011lexicon}.

Advancements in Machine learning (ML) \cite{pang2002thumbs}, such as supervised techniques trained on large amounts of labeled sentiment datasets, offer another powerful avenue for SA. 
Hence, the exploration of semi-supervised methods in SA could leverage unlabelled data to address the challenge of data annotation and labeling \cite{vo2015target, hwang2021semi}. Deep learning approaches such as Convolutional Neural Networks (CNN) \cite{chen2015convolutional} have significantly advanced SA capabilities. However, SA on social media poses unique challenges compared to more traditional domains due to the informal and conversational nature of the text \cite{medhat2014sentiment, naseem2019dice}. 

\subsection{Code-Switching on Low-resource}

Code-switching, the practice of alternating between two or more languages or dialects within a conversation, is particularly prevalent in multilingual communities and has become increasingly visible on social media platforms \cite{poplack2000toward,scotton1993social, danet2007multilingual}. It presents unique challenges and opportunities for NLP \cite{barman2014code}. Most NLP research traditionally focuses on high-resource languages like English, leaving low-resource languages underrepresented \cite{strassel2016lorelei, adelani2021masakhaner}. This gap is more pronounced in African and code-switched languages due to linguistic variability  \cite{adelani2021masakhaner}. Therefore, high-resource language techniques may underperform on low-resource language data \cite{lewis2014ethnologue}. The study in \cite{lee2015emotion} emphasizes the importance of analyzing emotions in code-switching data.  
The use of Generative Pre-trained Transformers (GPT) to generate synthetic code-switched data has been proposed to address data scarcity \cite{terblanche2024prompting}. A recent survey \cite{winata2022decades} revealed that until October 2022, only a few papers from the ACL Anthology and ISCA Proceedings focused on code-switching research in African languages. 
For South African languages \cite{niesler2018first, niesler2008accent} 
the first dataset was presented in 2018.
Even though Swahili-English code-switching has been studied in a few papers \cite{piergallini2016word, otundo2022intonation}, no datasets are available.

\subsection{ Transformer-based Pretrained Models}
Transformer-based architectures \cite{vaswani2017attention}, such as BERT  \cite{devlin2018bert}, have gained popularity owing to their effectiveness in learning general representations using large unlabelled datasets \cite{matthew2018peters} that can further be fine-tuned for downstream tasks \cite{gururangan2020don, bhattacharjee2020bert}. Hence, it has become the foundation for many NLP tasks \cite{bhattacharjee2020bert}.

\begin{figure}[ht!]
    \centering
    \includegraphics[width=1.0\columnwidth]{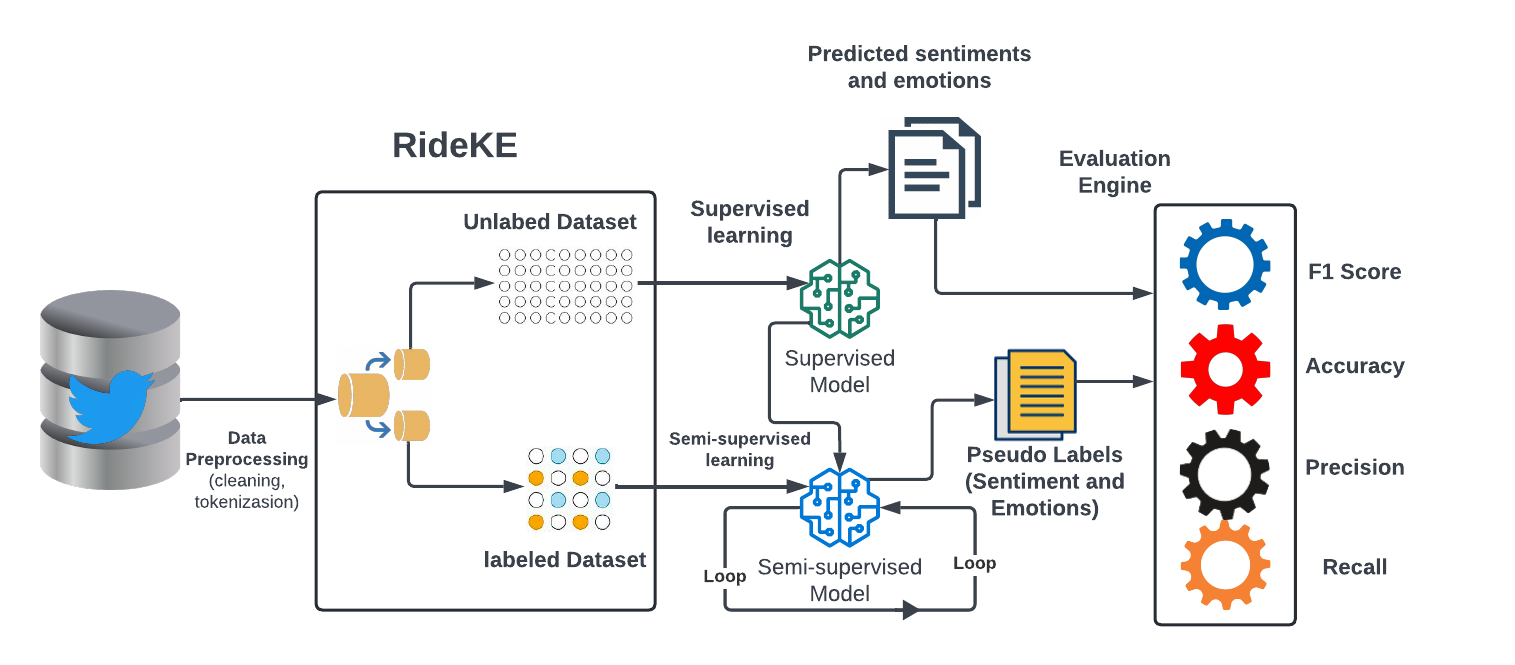}
    \caption{ \textbf{Methodology:} \textit{Overview of the RideKE sentiment and emotion analysis framework. Unlabeled and labeled datasets are preprocessed and used to train supervised and semi-supervised models for sentiment and emotion prediction. The semi-supervised learning loop generates pseudo labels for evaluation of performance.}}
    \label{fig:rideke}
\end{figure}

Pretrained language models are trained on large, diverse datasets \cite{raffel2020exploring}. For example, RoBERTa \cite{liu2019roberta} was pretrained on over 160GB of uncompressed text, from BOOKCORPUS \cite{zhu2015aligning} and CommonCrawl English dataset \cite{commoncrawlCommonCrawl}. These models learn representations that perform well across various tasks, handling datasets of different sizes from diverse sources while remaining easily understandable \cite{wang2019superglue}. Examples of a few applications in low-resource include improving speech recognition accuracy (ASR) \cite{olatunji2023afrispeech}, machine translation (MT)  \cite{wang2024afrimte} and SA \cite{muhammad2023afrisenti}. 

\section{Methods and Datasets}
\label{sxn:meth}
\subsection{Overview of RideKE Dataset}

RideKE dataset. as shown in Table~\ref{tab:tweets_sentiment_emotion} and ~\ref{tab:codeswitched}, includes a blend of Kenyan-accented English, approx. (70\%), with a minority mix of Swahili and Sheng (30\%). 
The dataset includes a total of 29,623 
entries across 12 distinct columns. See Table~\ref{tab:tweet_data_analysis} 
in the Appendix.

\subsection{Data Collection}
We used a systematic scraping process using the snscrape python library \footnote{\url{https://pypi.org/project/snscrape/l}} which allows for querying and retrieving tweets based on specified criteria. We targeted three keyword search terms—\#UBER-Kenya, \#BOLT-kenya, and \#LITTLECAB, from January 2017 to April 2023, capturing not only the tweet texts but also other essential metadata such as user engagement metrics (likes, retweets, replies), user account details (followers, following, tweet counts), and relational markers (hashtags, user mentions). Initially, the data was in a dictionary format but it was later converted to DataFrame using pandas and preserved in a CSV format to ensure reproducibility.

\subsubsection{Geo-based data collection}
The tweet's location metadata was crucial in determining the regional focus of our study. We referenced Kenya's location as shown in Table~\ref{table:tweet_counts_by_location}. To ensure uniformity, we used a simple yet effective keyword filtering normalization technique to address location inconsistencies as shown by the diverse representations of Nairobi in the dataset shown in Table~\ref{table:tweet_counts_by_location}.
To isolate the relevant tweets, we applied a filter on the \texttt{user\_location} field to include only locations mentioning \texttt{Kenya} and discard entries with missing data and all those with no location. 
We assessed the frequency distribution of different locations using value count function.

\begin{table}[ht]
\centering
\small 
\begin{tabular}{lr}
\toprule
\textbf{Location} & \textbf{Tweet Count} \\
\midrule
Kenya                      & 18974 \\
Nairobi, Kenya             & 11960 \\
\textit{Not specified}     & 10868 \\
Nairobi                    & 4776 \\
Nairobi, Kenya             & 620 \\
nairobery                  & 1 \\
Africa, Nairobi Kenya      & 1 \\
Mt. Meru                   & 1 \\
3rd Parklands              & 1 \\
New Jersey                 & 1 \\
\bottomrule
\end{tabular}
\caption { \textbf{Tweet Counts by location:}
\textit{We only included locations mentioning Kenya}
\label{table:tweet_counts_by_location}}
\end{table}

\subsection{Language Detection}
We used \texttt{langdetect} \footnote{\url{https://pypi.org/project/langdetect/l}} Python library to detect languages within text. It revealed diverse languages, \texttt{English}  being the most prevalent, then \texttt{Indonesian}, \texttt{Swahili} and others as shown in Table~\ref{tab:my_label}. For the Sheng language, native speakers manually detected the language. We only kept English (code-switched) for our analysis.

\subsection{Data Preprocessing}

Tweets often feature slang, abbreviations, and non-alphanumeric characters such as hashtags and emojis, contributing to the data's unstructured nature \cite{adebara2022towards}.  We implemented a refined text preprocessing pipeline to enhance data consistency and accurate analysis. The pipeline standardizes data by converting text to strings, trimming whitespace, lowering case, and expanding contractions to preserve semantic integrity. The text is then normalized by reducing repeated characters, removing punctuation, newlines, and tabs, and then tokenizing. 

\subsection{Data Annotation}
Inspired by \cite{raffel2020exploring} established guidelines, we created a set of annotation guidelines for emotion annotations to ensure a standardized and high-quality approach in our labeling efforts, as shown in Table~\ref{tab:emotion_guidelines}. We added a 'frustration' label and used 'happy' instead of 'joy.' For the sentiment annotation, we adhered to the established annotation framework detailed by \cite{mohammad2016practical}. However, human annotation is time-consuming and costly. We employed two Kenyan volunteer annotators fluent in English, Swahili, and Sheng. One holds a bachelor's degree in political science and the other in computer science. They received a small token of appreciation for their efforts. We ensured the annotator's comprehension of the task. Two annotators labeled the same dataset entries to enhance quality. Each labeled 1,554 tweets with sentiment labels (positive, negative, neutral) and emotion labels 
(sadness, happy, love, anger, fear, surprise, frustration, and neutral).

\subsubsection{Annotation Quality Control}
 We used Cohen's Kappa \cite{artstein2017inter}\footnote{\url{https://github.com/zyocum/cohens_kappa}} as our primary metric for assessing the level of inter-annotator agreement between the two annotators. It is perfect for categorical items, such as sentiment and emotion labels. Cohen’s Kappa provides a means to compute an inter-rater agreement score that accounts for the probability of random agreement:

\begin{equation}
\kappa = \frac{P_o - P_e}{1 - P_e}
\end{equation}
where $P_o$ is the observed agreement, and $P_e$ is the expected agreement by chance. 

To assign the final sentiment and emotion label to each tweet, we employed a majority voting method \cite{davani2022dealing} to determine the final label of the tweet \cite{mohammad2022ethics}. Instances of complete disagreement among annotators were resolved by involving a lead annotator and applying a majority rule rather than omitting them from the dataset. We found a Cohen's Kappa coefficient of 0.60 for sentiment classification tasks. Cohen's Kappa score for the emotion annotations is approximately 0.67, which indicates a substantial level of agreement beyond chance and suggests a good degree of consistency in their annotations.

\subsubsection{Data Splits}
The dataset was split into three sets (A, B, and C) as shown in the dataset division Table~\ref{table:dataset_division}. We used ChatGPT \cite{brown2020language} for automatic labeling to augment the training dataset and increase training labels since we had only two human annotators. Set A provided Ground truth labels for initial supervised training. Set B is the test dataset that is manually annotated by human annotators. Set C represented the unlabelled dataset Used in a semi-supervised training loop, with empty rows and duplicates removed, labels standardized and encoded.

\begin{table}[ht]
\centering
\small
\begin{tabular}{c l l}
\toprule
\textbf{Set} & \textbf{Description} & \textbf{Details} \\ \midrule
Set A & 553 human, 636 ChatGPT & Supervised Train \\ 
Set B & 2,000 human & Testing \\ 
Set C & 27,090 unlabelled & Semi-supervised \\ 
\bottomrule
\end{tabular}
\caption{Dataset Division}
\label{table:dataset_division}
\end{table}

\subsection{Semi-supervised Learning Phase}
Semi-supervised learning (SSL) offers a framework for utilizing large amounts of unlabelled data when obtaining labels is expensive \cite{chapelle2006introduction, learning2006semi} as applied to our case. Research shows SSL improves performance on different machine learning tasks such as text classification and machine translation \cite{najafi2019robustness}. SSL connects supervised and unsupervised learning by utilizing a small fraction of labelled data alongside a larger pool of unlabeled data to improve learning accuracy. SSL has been widely studied to show effectiveness for a wide range of low-resource applications, such as in text-to-speech synthesis (TTS) \cite{saeki2023virtuoso}, speech recognition \cite{du2023semi,thomas2013deep}, machine translation\cite{pham2023semi,singh2022low}, POS-Taggers \cite{garrette2013real}, and sentiment classification \cite{gupta2018semi}. Our work extends the application of SSL to sentiment and emotion classification tasks. We seek to mitigate this limitation by leveraging labeled and unlabeled 
data to train pretrained models. We used accuracy, precision, recall, and F1 scores to evaluate the models' performance.

\begin{table*}[!ht]
    \centering
    \small
    \resizebox{1.0\textwidth}{!}{%
    \begin{tabular}{l|cccc|cccc}
    \toprule
     & \multicolumn{4}{c|}{Sentiment} & \multicolumn{4}{c}{Emotions} \\
    Model & Accuracy & Precision & Recall & F1-Score & Accuracy & Precision & Recall & F1-Score \\
    \midrule
    DistilBERT supervised & 0.578 & 0.598 & 0.629 & 0.546 & 0.598 & 0.334 & 0.315 & 0.310 \\
    DistilBERT semi-supervised & 0.553 & 0.585 & 0.598 & 0.516 & 0.544 & 0.264 & 0.266 & 0.252 \\
    mBERT supervised & 0.638 & 0.621 & 0.663 & 0.596 & 0.592 & 0.253 & 0.298 & 0.265 \\
    mBERT semi-supervised & 0.635 & 0.622 & 0.661 & 0.598 & 0.594 & 0.297 & 0.317 & 0.297 \\
    XLM-R supervised & 0.692 & 0.665 & 0.723 & 0.661 & 0.658 & 0.343 & 0.267 & 0.258 \\
    XLM-R semi-supervised & 0.672 & 0.644 & 0.702 & 0.641 & 0.620 & 0.334 & 0.248 & 0.230 \\
    AfriBERTa large supervised & 0.398 & 0.500 & 0.479 & 0.358 & 0.604 & 0.163 & 0.191 & 0.157 \\
    AfriBERTa semi-supervised & 0.413 & 0.534 & 0.491 & 0.366 & 0.556 & 0.145 & 0.177 & 0.142 \\
    \bottomrule
    \end{tabular}}
    \caption{\textbf{Model Performance Evaluation on Sentiment and Emotion Analysis Tasks.} \textit{Performance evaluation of supervised and semi-supervised training for sentiment and emotion analysis across models. Results represent averages over multiple runs.}}
    \label{tab:model_performance_sentiment_emotion}
\end{table*}

\begin{table*}[!ht]
    \centering
    \small
     \resizebox{1.0\textwidth}{!}{%
    \begin{tabular}{l|ccc|ccc|ccc}
    \toprule
     & \multicolumn{3}{c|}{Negative} & \multicolumn{3}{c|}{Neutral} & \multicolumn{3}{c}{Positive} \\
    Model & Precision & Recall & F1 & Precision & Recall & F1 & Precision & Recall & F1 \\
    \midrule
    DistilBERT supervised & 0.920 & 0.385 & 0.543 & 0.284 & 0.635 & 0.392 & - & - & - \\
    DistilBERT semi-supervised & 0.901 & 0.325 & 0.478 & 0.268 & 0.604 & 0.371 & - & - & - \\
    mBERT supervised & 0.906 & 0.467 & 0.616 & 0.330 & 0.587 & 0.423 & - & - & - \\
    mBERT semi-supervised & 0.873 & 0.443 & 0.588 & 0.363 & 0.628 & 0.460 & - & - & - \\
    XLM-R supervised & 0.921 & 0.563 & 0.699 & 0.417 & 0.714 & 0.526 & - & - & - \\
    XLM-R semi-supervised & 0.850 & 0.524 & 0.648 & 0.392 & 0.712 & 0.506 & 0.691 & 0.871 & 0.771 \\
    AfriBERTa large supervised & 0.794 & 0.100 & 0.178 & 0.144 & 0.492 & 0.223 & - & - & - \\
    AfriBERTa semi-supervised & 0.874 & 0.096 & 0.174 & 0.171 & 0.560 & 0.261 & 0.558 & 0.817 & 0.663 \\
    \bottomrule
    \end{tabular}}
    \caption{\textbf{Model Performance Evaluation on Sentiment classification Tasks Labels.} \textit{Performance evaluation for Negative, Neutral, and Positive sentiments across various models. A dash (-) indicates missing values, i.e., the models did not predict all positive sentiment instances. The results represent averages over multiple runs.}}
    \label{tab:model_performance_sentiment}
\end{table*}

 \section{Experiments}
\label{sxn:exp}
\subsection{Models and Architecture}
We evaluate four transformer-based models in our experiments: \textbf{DistilBERT} \cite{sanh2019distilbert}, a smaller and faster version of BERT;  \textbf{mBERT} \cite{devlin2018bert}, a multilingual version of BERT trained on 104 languages; \textbf{XLM-RoBERTa} \cite{conneau2019unsupervised}, a multilingual model trained on 100 languages with improved performance; and \textbf{AfriBERTa large} \cite{ogueji2021small}, a model specifically designed for African languages to address the unique linguistic challenges in this region. Each model was trained on supervised and semi-supervised learning on sentiment and emotion classification tasks. The initial supervised training and subsequent semi-supervised fine-tuning were conducted separately for each model.

\subsection{Experimental Setup }
\subsubsection{Supervised Learning Phase}
In supervised training, we utilized the human-annotated, well-curated labeled dataset. We used batches ranging from 16 to 64 depending on the model sizes, optimizing for computational efficiency. A combined categorical cross-entropy loss shown in Figure \ref{fig:TrainingLosses} function, with equal weighting for sentiment and emotion tasks, guided the model toward effective multitasking. We applied a dropout rate of 0.1 for each model to prevent overfitting and enhance generalization. 
We employed the Adam optimizer, with a learning rate $1e-5$ through 10 epochs of training and monitoring. 
Initially, the four transformer-based models were fine-tuned on a dataset with 1,189 labeled tweets. We then evaluated the model.

\subsubsection{Semi-supervised Learning Phase}

Our goal in using SSL is to leverage the vast, unlabeled datasets to mitigate the high cost of human annotations. Following an initial supervised learning phase, each transformer-based model underwent a semi-supervised training loop. In this loop, the models dynamically labeled the unlabeled dataset based on their predictions, generating a pseudo-labeled dataset. We employed a dynamic threshold, set at the 75th percentile of the models' probability predictions across all classes for each batch, to ensure only high-confidence predictions were used for labeling. Samples with predictions below this threshold were excluded to minimize the inclusion of erroneous labels in the training data.

We extended the semi-supervised training loop over 4 epochs, a duration we empirically selected to refine the models' generalization capabilities without causing performance degradation due to overtraining, as indicated by either worsening or plateauing loss. We carefully chose the hyperparameters to ensure optimal training dynamics and model performance.

\begin{figure}[ht!]
\centering
\begin{subfigure}{0.49\textwidth}
  \centering
  \includegraphics[width=\linewidth]{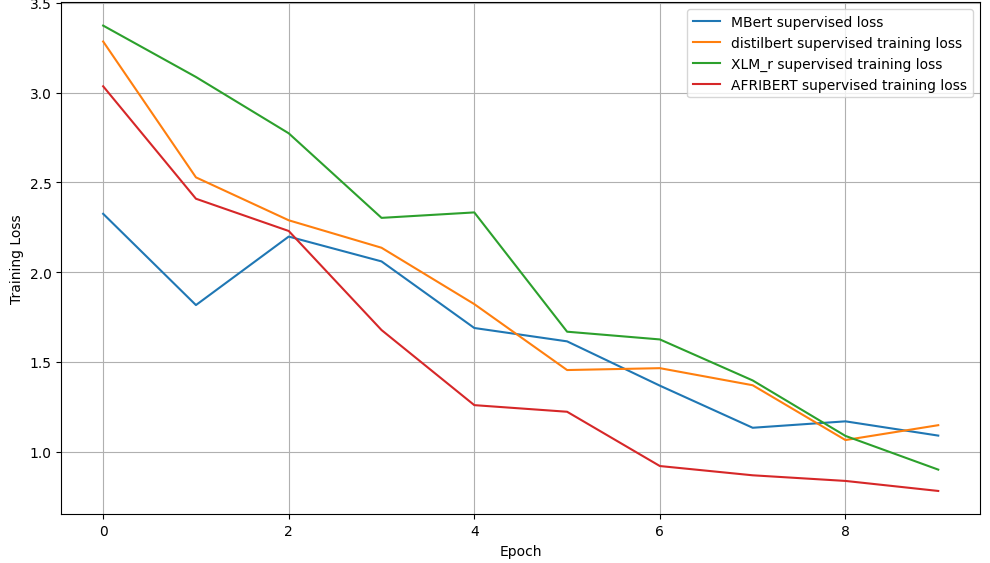}
  \caption{Supervised Loss}
  \label{fig:SupervisedTrainingLoss}
\end{subfigure}
\begin{subfigure}{0.49\textwidth}
  \centering
  \includegraphics[width=\linewidth]{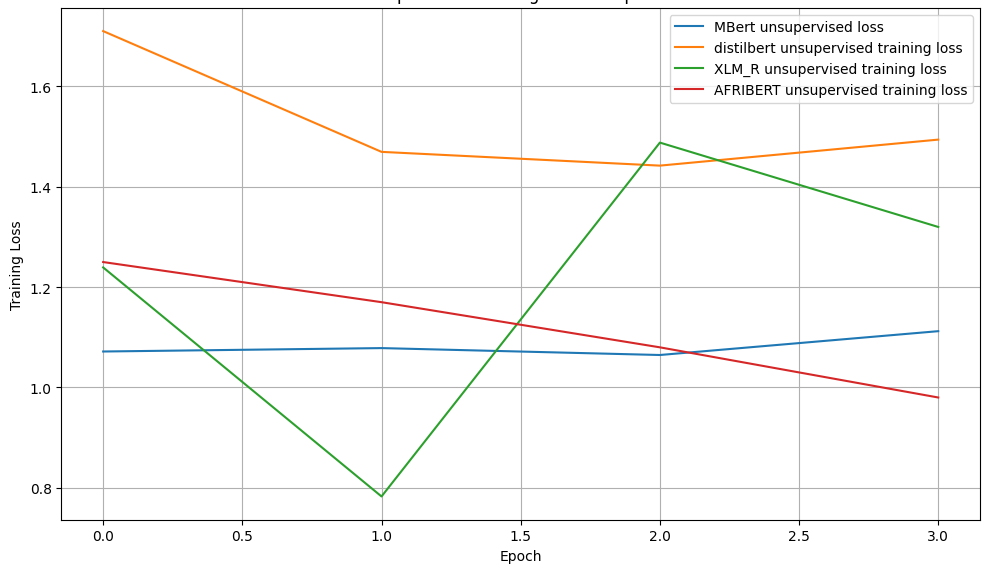}
  \caption{Semi-supervised Loss}
  \label{fig:SemiSupervisedTrainingLoss}
\end{subfigure}
\caption{\textbf{Training loss} \textit{ (a) supervised and (b) semi-supervised learning.}}
\label{fig:TrainingLosses}
\end{figure}

We set the learning rate at 1e-5 and dynamically adjusted it using a learning rate scheduler during training to optimize generalization and reduce overfitting. The batch size varied between 16 and 64, depending on the specific transformer model, to ensure computational efficiency. We used a combined loss function shown in Figure \ref{fig:TrainingLosses} for sentiment and emotion analysis and applied a dropout rate of 0.1 to prevent overfitting. We employed the Adam optimizer with a learning rate of 1e-5 and no weight decay. 

\begin{table*}[!ht]
    \centering
    \small
    \resizebox{1.0\textwidth}{!}{%
    \begin{tabular}{l|ccc|ccc|ccc}
    \toprule
     & \multicolumn{3}{c|}{Neutral} & \multicolumn{3}{c|}{Frustration} & \multicolumn{3}{c}{Happy} \\
    Metrics & Precision & Recall & F1-score & Precision & Recall & F1-score & Precision & Recall & F1-score \\
    \midrule
    Distilbert\_supervised\ & 0.130 & 0.176 & 0.150 & 0.444 & 0.364 & 0.400 & 0.000 & 0.000 & 0.000 \\
    Distilbert\_semi\_supervised & 0.141 & 0.121 & 0.131 & 0.132 & 0.227 & 0.167 & 0.000 & 0.000 & 0.000 \\
    mBERT\_supervised\ & 0.043 & 0.059 & 0.050 & 0.000 & 0.000 & 0.000 & 0.000 & 0.000 & 0.000 \\
    mBERT\_semi\_supervised & 0.284 & 0.234 & 0.256 & 0.100 & 0.045 & 0.063 & 0.000 & 0.000 & 0.000 \\
    XLM\_R\_supervised\_training & 1.000 & 0.118 & 0.211 & 0.000 & 0.000 & 0.000 & 0.000 & 0.000 & 0.000 \\
    XML\_R\_semi\_supervised & 0.571 & 0.037 & 0.070 & 0.333 & 0.015 & 0.029 & 0.000 & 0.000 & 0.000 \\
    AfriBERTa\_large\_supervised & 0.000 & 0.000 & 0.000 & 0.000 & 0.000 & 0.000 & 0.000 & 0.000 & 0.000 \\
    AfriBERTa\_semi\_supervised & 0.000 & 0.000 & 0.000 & 0.000 & 0.000 & 0.000 & 0.000 & 0.000 & 0.000 \\
    \bottomrule
    \end{tabular}
    }
    \caption{\textbf{Model Performance Evaluation on Emotion classification Tasks.} \textit{Performance metrics of supervised and semi-supervised learning for (Neutral, Frustration, and Happy) emotion analysis across models. Showing poor performance of happy emotions.}}
    \label{tab:model_performance_sentiment_emotion_table1}
\end{table*}

\begin{table*}[!ht]
    \centering
    \small
    \resizebox{1.0\textwidth}{!}{%
    \begin{tabular}{l|ccc|ccc|ccc}
    \toprule
     & \multicolumn{3}{c|}{Anger} & \multicolumn{3}{c|}{Love} & \multicolumn{3}{c}{Fear} \\
    Model & Precision & Recall & F1-Score & Precision & Recall & F1-Score & Precision & Recall & F1-Score \\
    \midrule
    Distilbert\_supervised\ & 0.517 & 0.861 & 0.646 & 0.333 & 0.222 & 0.267 & 0.000 & 0.000 & 0.000 \\
    Distilbert\_semi\_supervised & 0.445 & 0.833 & 0.580 & 0.357 & 0.212 & 0.266 & 0.000 & 0.000 & 0.000 \\
    mBERT\_supervised\ & 0.524 & 0.795 & 0.632 & 0.408 & 0.444 & 0.426 & 0.000 & 0.000 & 0.000 \\
    mBERT\_semi\_supervised & 0.487 & 0.838 & 0.616 & 0.438 & 0.430 & 0.434 & 0.000 & 0.000 & 0.000 \\
    XLM\_R\_supervised\ & 0.553 & 0.943 & 0.697 & 0.489 & 0.489 & 0.489 & 0.000 & 0.000 & 0.000 \\
    XML\_R\_semi\_supervised & 0.506 & 0.918 & 0.652 & 0.427 & 0.461 & 0.443 & 0.000 & 0.000 & 0.000 \\
    AfriBERTa\_large\_supervised & 0.484 & 0.975 & 0.647 & 0.250 & 0.022 & 0.041 & 0.000 & 0.000 & 0.000 \\
    AfriBERTa\_semi\_supervised & 0.417 & 0.920 & 0.574 & 0.182 & 0.012 & 0.023 & 0.000 & 0.000 & 0.000 \\
    \bottomrule
    \end{tabular}}
    \caption{\textbf{Model Performance Evaluation on Emotion Classification Tasks.} \textit{Performance metrics of supervised and semi-supervised training for (Anger, Love, and Fear) emotion analysis across models. The model performed poorly on Fear emotions.}}
    \label{tab:model_performance_emotion_analysis_table2}
\end{table*}

\begin{table*}[!ht]
    \centering
    \small
    \resizebox{1.0\textwidth}{!}{%
    \begin{tabular}{l|ccc|ccc|ccc}
    \toprule
     & \multicolumn{3}{c|}{Sadness} & \multicolumn{3}{c|}{Empathy} & \multicolumn{3}{c}{Surprise} \\
    Model & Precision & Recall & F1-Score & Precision & Recall & F1-Score & Precision & Recall & F1-Score \\
    \midrule
    Distilbert\_supervised\ & 0.500 & 0.222 & 0.308 & 0.833 & 0.652 & 0.732 & 0.250 & 0.333 & 0.286 \\
    Distilbert\_semi\_supervised & 0.100 & 0.083 & 0.091 & 0.844 & 0.580 & 0.688 & 0.360 & 0.337 & 0.348 \\
    mBERT\_supervised\ & 0.200 & 0.222 & 0.211 & 0.865 & 0.660 & 0.749 & 0.237 & 0.500 & 0.321 \\
    mBERT\_semi\_supervised & 0.129 & 0.167 & 0.145 & 0.855 & 0.621 & 0.720 & 0.377 & 0.516 & 0.436 \\
    XLM\_R\_supervised\ & 0.250 & 0.111 & 0.154 & 0.791 & 0.747 & 0.768 & 0.000 & 0.000 & 0.000 \\
    XML\_R\_semi\_supervised & 0.154 & 0.083 & 0.108 & 0.767 & 0.701 & 0.733 & 0.250 & 0.021 & 0.039 \\
    AfriBERTa\_large\_supervised & 0.000 & 0.000 & 0.000 & 0.731 & 0.719 & 0.725 & 0.000 & 0.000 & 0.000 \\
    AfriBERTa\_semi\_supervised & 0.000 & 0.000 & 0.000 & 0.706 & 0.659 & 0.682 & 0.000 & 0.000 & 0.000 \\
    \bottomrule
    \end{tabular}}
    \caption{\textbf{Model Performance Evaluation on Emotion classification Tasks.} \textit{Performance metrics of supervised and semi-supervised training methods for emotion (Sadness, Empathy, and Surprise) analysis across various models. Showing outstanding performance on  Empathy emotions.}}
    \label{tab:model_performance_emotion_analysis_table3}
\end{table*}

\section{Results and Discussions}
\label{sxn:res}

\subsection{Sentiment Analysis}
Table~\ref{tab:model_performance_sentiment_emotion} summarizes the performance of all models on sentiment analysis. XLM-R supervised achieves the highest overall performance with an accuracy of 62.5\%  and an F1-score of 66.7\%. This is followed closely with semi-supervised XLM-R, which has an accuracy of 62.1\% and an F1-score of 68.3\%. However, DistilBERT supervised performance falls behind with an accuracy of 57.8\%  and an F1-score of 54.6\%. On the other hand, mBERT models show consistency between supervised and semi-supervised training, maintaining average F1-scores of 59.8\% and 59.6\%, respectively. AfriBERTa models struggled, with the supervised learning achieving an F1-score of 35.8\%, and overall poorest performance across all metrics.

The detailed performance metrics for negative, neutral, and positive sentiment classification are presented in Table~\ref{tab:model_performance_sentiment}.  For the negative sentiment, the supervised XLM-R achieves a high F1-score of 69.9\%, unlike the semi-supervised AfriBERTa, which has the worst F1-score of 17.4\%. In neutral sentiment classification, the supervised XLM-R again excels with an F1-score of 52.6\%. For the positive sentiment, the semi-supervised XLM-R stands out with an exceptional F1-score of 77.1\%, and the semi-supervised AfriBERTa shows robust performance with an F1-score of 66.3\%. 

\begin{figure*}[h]
    \centering
    \begin{subfigure}[t]{0.46\textwidth}
        \centering
        \includegraphics[width=\textwidth]{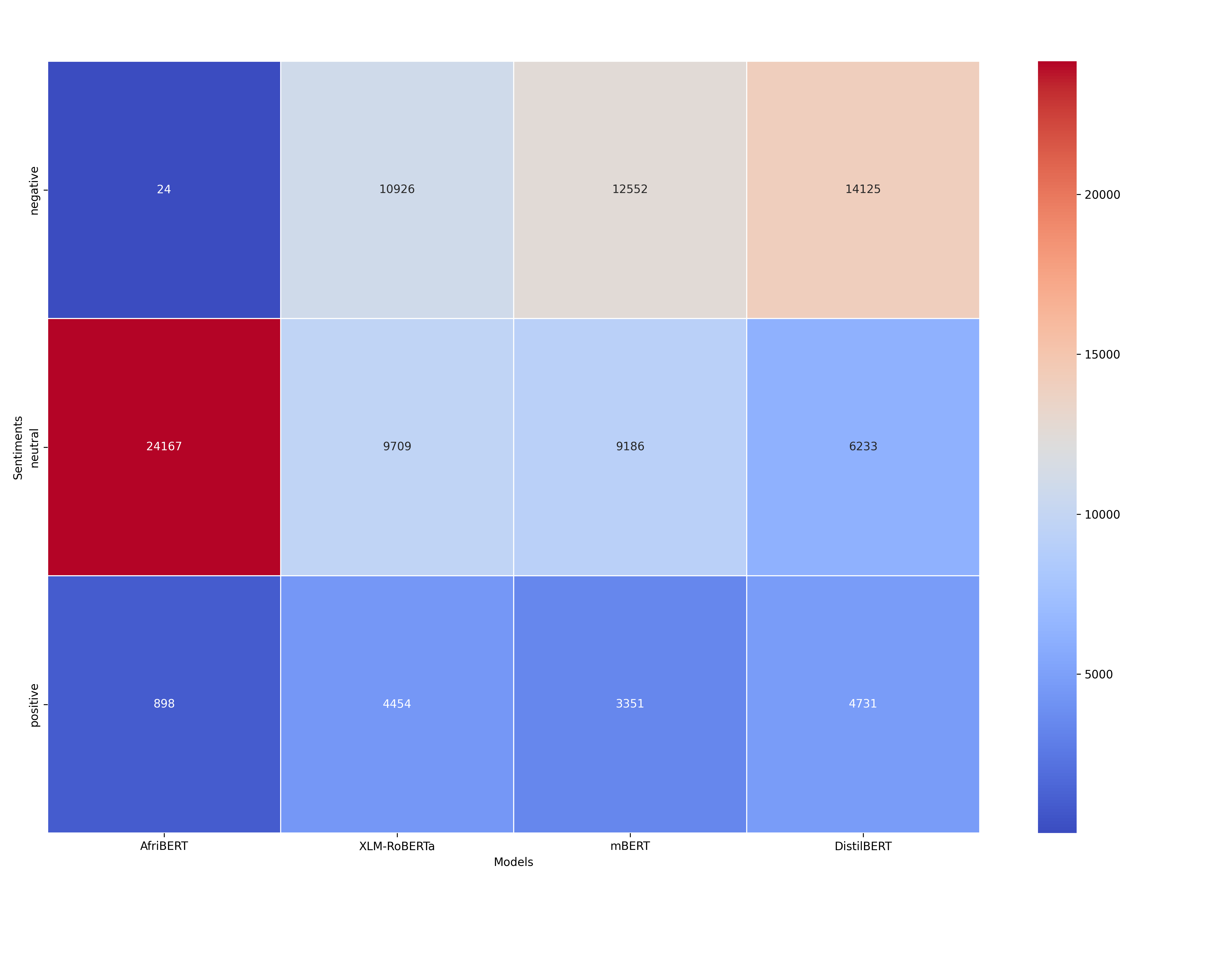}
        \caption{Sentiment Prediction Comparison Across Models}
    \end{subfigure}
    \hfill
    \begin{subfigure}[t]{0.48\textwidth}
        \centering
        \includegraphics[width=\textwidth]{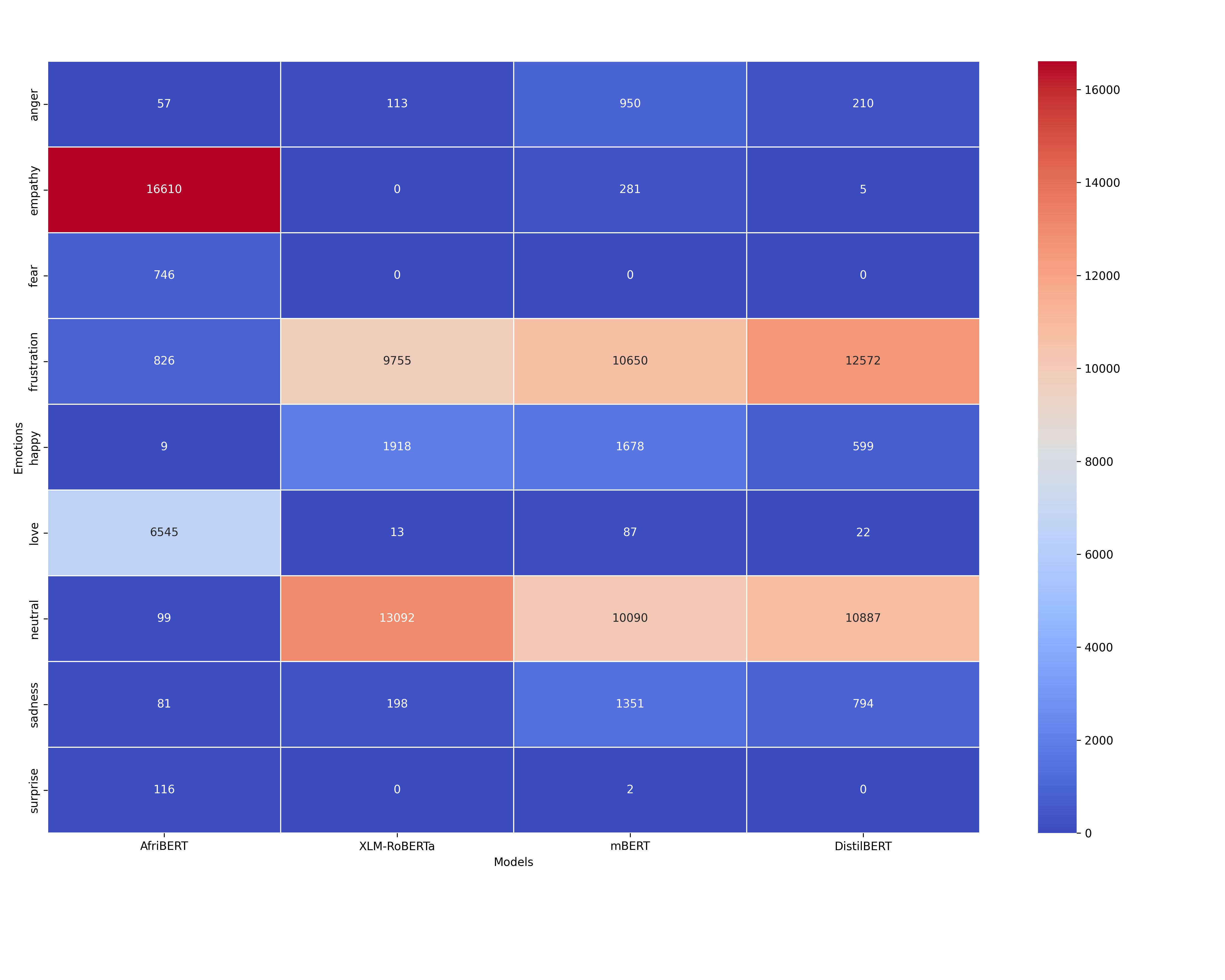}
        \caption{Emotion Prediction Comparison Across Models}
    \end{subfigure}    
    \caption{\textit{Heatmaps comparing sentiment and emotion predictions across different models. AfriBERT model most frequently predicts neutral sentiment and shows the highest sensitivity for empathy emotions.}}
    \label{fig:heatmaps}
\end{figure*}

\subsection{Emotion Analysis}
Table~\ref{tab:model_performance_sentiment_emotion} summarizes the performance of all models on emotion analysis. The models generally show lower performance than sentiment analysis. Since emotions are complex \cite{mohammad2017word}. The supervised DistilBERT achieves the highest F1-score of 31\%, followed by mBERT semi-supervised, with an F1-score of 29.7\%.

Table~\ref{tab:model_performance_sentiment_emotion_table1} shows  performance for emotion classification across neutral, frustration, and happy. DistilBERT supervised leads in frustration with an F1-score of 40\%. All models perform poorly on happy emotion classification. In Table~\ref{tab:model_performance_emotion_analysis_table2}, XLM-R supervised leads for anger and love emotions with F1-scores of 69.7\% and 48.9\%, respectively, but all models struggle with fear emotion. Table~\ref{tab:model_performance_emotion_analysis_table3} shows low performance for sadness and surprise  but outstanding performance for empathy with XLM-R supervised, leading with an F1-score of 76.8\%.

\subsection{Pretrained Models performance}
As shown in Figure~\ref{fig:heatmaps}, XLM-R, particularly in its supervised form, consistently outperforms other models across sentiment and emotion analysis tasks. mBERT also performs reliably well in sentiment analysis and some emotion classifications. DistilBERT, while efficient, has limitations in handling a range of emotions. AfriBERTa shows lower performance across most metrics than other models. Despite being tailored to African languages, AfriBERTa models do not perform as well in sentiment and even worse in emotion analysis.

\subsection{Semi-Supervised Performance Analysis}
The detailed analysis of SSL models reveals mixed outcomes, with clear performance enhancements in certain models and tasks, particularly in sentiment analysis. For example, mBERT's semi-supervised version slightly improved sentiment analysis with an F1-score of 59.8\% compared to 59.6\% for supervised version. In emotion analysis, mBERT's semi-supervised version outperformed its supervised counterpart with an F1-score of 29.7\% versus 26.5\%. The semi-supervised AfriBERTa achieved an F1-score of 36.6\% in sentiment analysis, marginally higher than the supervised version's 35.8\%, and scored 15.7\% compared to 14.2\% in emotion task.

\section{Limitations}
We acknowledge the subjective nature of sentiment and emotion analysis, which can be influenced by label bias, leading to inconsistencies in labeled data. We will publicly share our dataset to address this issue and facilitate further study on label bias and annotator disagreement. 
Secondly, the cost of obtaining labeled datasets, particularly from native speakers, can be challenging. Transformer models, SOTA for sentiment and emotion analysis, require large data and computational resources, which is still challenging in low-resource setting. Lastly, We recognize the ethical considerations of LLM use.

\section{Conclusions and Future Work}
\label{sec:conclusion}
We presented RideKE, a code-switched dataset from Twitter, with sentiment and emotion labels partially annotated for Kenyan-accented English mixed with Swahili and Sheng. Our semi-supervised learning shows mixed results, with clear performance enhancements in certain models and tasks, particularly in sentiment analysis, suggesting its potential to generally enhance model performance. We highlight the benefits of semi-supervised learning in improving model performance and reducing data annotation costs.

In the future, we aim to further enhance model performance by expanding the pool of human-labeled datasets, using other semi-supervised approaches, utilizing techniques like few-shot learning, and experimenting with different model architectures and hyperparameters tuning.

\section*{Acknowledgments}
We thank the volunteer annotators who dedicated their time and expertise to this project, which would not have succeeded without their commitment. 



\bibliography{anthology, custom}

\begin{thebibliography}{81}
\expandafter\ifx\csname natexlab\endcsname\relax\def\natexlab#1{#1}\fi

\bibitem[{Adebara and Abdul-Mageed(2022)}]{adebara2022towards}
Ife Adebara and Muhammad Abdul-Mageed. 2022.
\newblock \href {https://aclanthology.org/2022.acl-long.265/} {Towards afrocentric nlp for african languages: Where we are and where we can go}.
\newblock \emph{arXiv preprint arXiv:2203.08351}.

\bibitem[{Adelani et~al.(2021)Adelani, Abbott, Neubig, D’souza, Kreutzer, Lignos, Palen-Michel, Buzaaba, Rijhwani, Ruder et~al.}]{adelani2021masakhaner}
David~Ifeoluwa Adelani, Jade Abbott, Graham Neubig, Daniel D’souza, Julia Kreutzer, Constantine Lignos, Chester Palen-Michel, Happy Buzaaba, Shruti Rijhwani, Sebastian Ruder, et~al. 2021.
\newblock \href {https://direct.mit.edu/tacl/article/doi/10.1162/tacl_a_00416/107614/MasakhaNER-Named-Entity-Recognition-for-African} {Masakhaner: Named entity recognition for african languages}.
\newblock \emph{Transactions of the Association for Computational Linguistics}, 9:1116--1131.

\bibitem[{Angel et~al.(2020)Angel, Aroyehun, Tamayo, and Gelbukh}]{angel2020nlp}
Jason Angel, Segun~Taofeek Aroyehun, Antonio Tamayo, and Alexander Gelbukh. 2020.
\newblock \href {https://arxiv.org/abs/2009.03397} {{NLP-CIC at SemEval}-2020 task 9: Analysing sentiment in code-switching language using a simple deep-learning classifier}.
\newblock \emph{arXiv preprint arXiv:2009.03397}.

\bibitem[{Artstein(2017)}]{artstein2017inter}
Ron Artstein. 2017.
\newblock \href {https://apps.dtic.mil/sti/trecms/pdf/AD1158943.pdf} {Inter-annotator agreement}.
\newblock \emph{Handbook of linguistic annotation}, pages 297--313.

\bibitem[{Aryal et~al.(2023)Aryal, Prioleau, and Aryal}]{aryal2023sentiment}
Saurav~K Aryal, Howard Prioleau, and Surakshya Aryal. 2023.
\newblock \href {https://arxiv.org/abs/2310.14120} {Sentiment analysis across multiple african languages: A current benchmark}.
\newblock \emph{arXiv preprint arXiv:2310.14120}.

\bibitem[{Aytu{\u{g}}(2018)}]{aytuug2018sentiment}
ONAN Aytu{\u{g}}. 2018.
\newblock \href {https://dergipark.org.tr/tr/download/article-file/465465} {Sentiment analysis on twitter based on ensemble of psychological and linguistic feature sets}.
\newblock \emph{Balkan Journal of Electrical and Computer Engineering}, 6(2):69--77.

\bibitem[{Baccianella et~al.(2010)Baccianella, Esuli, and Sebastiani}]{baccianella2010sentiwordnet}
Stefano Baccianella, Andrea Esuli, and Fabrizio Sebastiani. 2010.
\newblock \href {https://aclanthology.org/L10-1531/} {Sentiwordnet 3.0: An enhanced lexical resource for sentiment analysis and opinion mining}.
\newblock In \emph{Proceedings of the Seventh International Conference on Language Resources and Evaluation (LREC'10)}, pages 2200--2204.

\bibitem[{Banks-Wallace(2002)}]{banks2002talk}
JoAnne Banks-Wallace. 2002.
\newblock \href {https://journals.sagepub.com/doi/abs/10.1177/104973202129119892} {Talk that talk: Storytelling and analysis rooted in african american oral tradition}.
\newblock \emph{Qualitative health research}, 12(3):410--426.

\bibitem[{Barasa(2016)}]{barasa2016spoken}
Sandra Barasa. 2016.
\newblock \href {https://www.researchgate.net/publication/278963928_Spoken_Code-Switching_in_Written_Form_Manifestation_of_Code-Switching_in_Computer_Mediated_Communication} {Spoken code-switching in written form? manifestation of code-switching in computer mediated communication}.
\newblock \emph{Journal of Language Contact}, 9(1):49--70.

\bibitem[{Barman et~al.(2014)Barman, Das, Wagner, and Foster}]{barman2014code}
Utsab Barman, Amitava Das, Joachim Wagner, and Jennifer Foster. 2014.
\newblock \href {https://aclanthology.org/W14-3902.pdf} {Code mixing: A challenge for language identification in the language of social media}.
\newblock In \emph{Proceedings of the First Workshop on Computational Approaches to Code Switching}, pages 13--23.

\bibitem[{Bhattacharjee et~al.(2020)Bhattacharjee, Ballesteros, Anubhai, Muresan, Ma, Ladhak, and Al-Onaizan}]{bhattacharjee2020bert}
Kasturi Bhattacharjee, Miguel Ballesteros, Rishita Anubhai, Smaranda Muresan, Jie Ma, Faisal Ladhak, and Yaser Al-Onaizan. 2020.
\newblock \href {https://arxiv.org/abs/2010.14042} {To {BERT} or not to {BERT}: Comparing task-specific and task-agnostic semi-supervised approaches for sequence tagging}.
\newblock \emph{arXiv preprint arXiv:2010.14042}.

\bibitem[{Brown et~al.(2020)Brown, Mann, Ryder, Subbiah, Kaplan, Dhariwal, Neelakantan, Shyam, Sastry, Askell et~al.}]{brown2020language}
Tom Brown, Benjamin Mann, Nick Ryder, Melanie Subbiah, Jared~D Kaplan, Prafulla Dhariwal, Arvind Neelakantan, Pranav Shyam, Girish Sastry, Amanda Askell, et~al. 2020.
\newblock \href {https://arxiv.org/abs/2005.14165} {Language models are few-shot learners}.
\newblock \emph{Advances in Neural Information Processing Systems}, 33:1877--1901.

\bibitem[{Carter-Black(2007)}]{carter2007teaching}
Jan Carter-Black. 2007.
\newblock \href {https://www.tandfonline.com/doi/abs/10.5175/JSWE.2007.200400471} {Teaching cultural competence: An innovative strategy grounded in the universality of storytelling as depicted in african and african american storytelling traditions}.
\newblock \emph{Journal of Social Work Education}, 43(1):31--50.

\bibitem[{Chapelle et~al.(2006)Chapelle, Sch{\"o}lkopf, and Zien}]{chapelle2006introduction}
Olivier Chapelle, Bernhard Sch{\"o}lkopf, and Alexander Zien. 2006.
\newblock \href {https://direct.mit.edu/books/edited-volume/3824/chapter-abstract/125431/Introduction-to-Semi-Supervised-Learning?redirectedFrom=PDF} {\emph{Introduction to Semi-Supervised Learning}}.
\newblock MIT press.

\bibitem[{Chen(2015)}]{chen2015convolutional}
Yahui Chen. 2015.
\newblock \href {https://uwspace.uwaterloo.ca/items/42654efd-45e2-4c67-b906-158e7e349188} {Convolutional neural network for sentence classification}.
\newblock Master's thesis, University of Waterloo.

\bibitem[{Conneau et~al.(2019)Conneau, Khandelwal, Goyal, Chaudhary, Wenzek, Guzm{\'a}n, Grave, Ott, Zettlemoyer, and Stoyanov}]{conneau2019unsupervised}
Alexis Conneau, Kartikay Khandelwal, Naman Goyal, Vishrav Chaudhary, Guillaume Wenzek, Francisco Guzm{\'a}n, Edouard Grave, Myle Ott, Luke Zettlemoyer, and Veselin Stoyanov. 2019.
\newblock \href {https://arxiv.org/abs/1911.02116} {Unsupervised cross-lingual representation learning at scale}.
\newblock \emph{arXiv preprint arXiv:1911.02116}.

\bibitem[{Danet and Herring(2007)}]{danet2007multilingual}
Brenda Danet and Susan~C Herring. 2007.
\newblock \href {https://academic.oup.com/book/32471} {\emph{The multilingual Internet: Language, culture, and communication online}}.
\newblock Oxford University Press.

\bibitem[{Das and Chen(2001)}]{das2001yahoo}
Sanjiv~Ranjan Das and Mike~Y Chen. 2001.
\newblock \href {https://pubsonline.informs.org/doi/10.1287/mnsc.1070.0704} {Yahoo! for amazon: Sentiment parsing from small talk on the web}.
\newblock \emph{For Amazon: Sentiment Parsing from Small Talk on the Web (August 5, 2001). EFA}.

\bibitem[{Davani et~al.(2022)Davani, D{\'\i}az, and Prabhakaran}]{davani2022dealing}
Aida~Mostafazadeh Davani, Mark D{\'\i}az, and Vinodkumar Prabhakaran. 2022.
\newblock \href {https://direct.mit.edu/tacl/article/doi/10.1162/tacl_a_00449/109286/Dealing-with-Disagreements-Looking-Beyond-the} {Dealing with disagreements: Looking beyond the majority vote in subjective annotations}.
\newblock \emph{Transactions of the Association for Computational Linguistics}, 10:92--110.

\bibitem[{Dave et~al.(2003)Dave, Lawrence, and Pennock}]{dave2003mining}
Kushal Dave, Steve Lawrence, and David~M Pennock. 2003.
\newblock \href {https://dl.acm.org/doi/10.1145/775152.775226} {Mining the peanut gallery: Opinion extraction and semantic classification of product reviews}.
\newblock In \emph{Proceedings of the 12th International Conference on World Wide Web}, pages 519--528.

\bibitem[{Devlin et~al.(2018)Devlin, Chang, Lee, and Toutanova}]{devlin2018bert}
Jacob Devlin, Ming-Wei Chang, Kenton Lee, and Kristina Toutanova. 2018.
\newblock \href {https://arxiv.org/abs/1810.04805} {Bert: Pre-training of deep bidirectional transformers for language understanding}.
\newblock \emph{arXiv preprint arXiv:1810.04805}.

\bibitem[{Du et~al.(2023)Du, Zhang, Fang, Wu, and Yang}]{du2023semi}
Ye-Qian Du, Jie Zhang, Xin Fang, Ming-Hui Wu, and Zhou-Wang Yang. 2023.
\newblock \href {https://ieeexplore.ieee.org/document/10246368} {A semi-supervised complementary joint training approach for low-resource speech recognition}.
\newblock \emph{IEEE/ACM Transactions on Audio, Speech, and Language Processing}.

\bibitem[{Dwivedi(2014)}]{dwivedi2014linguistic}
Amitabh~Vikram Dwivedi. 2014.
\newblock \href {https://laghana.org/gjl/index.php/gjl/article/view/20} {Linguistic realities in {K}enya: A preliminary survey}.
\newblock \emph{Ghana Journal of Linguistics}, 3(2):27--34.

\bibitem[{Garrette et~al.(2013)Garrette, Mielens, and Baldridge}]{garrette2013real}
Dan Garrette, Jason Mielens, and Jason Baldridge. 2013.
\newblock \href {https://aclanthology.org/P13-1057/} {Real-world semi-supervised learning of {POS}-taggers for low-resource languages}.
\newblock In \emph{Proceedings of the 51st Annual Meeting of the Association for Computational Linguistics (Volume 1: Long Papers)}, pages 583--592.

\bibitem[{Gupta et~al.(2018)Gupta, Sahu, Espy-Wilson, and Narayanan}]{gupta2018semi}
Rahul Gupta, Saurabh Sahu, Carol Espy-Wilson, and Shrikanth Narayanan. 2018.
\newblock \href {https://ieeexplore.ieee.org/document/8461414} {Semi-supervised and transfer learning approaches for low resource sentiment classification}.
\newblock In \emph{2018 IEEE international conference on acoustics, speech and signal processing (ICASSP)}, pages 5109--5113. IEEE.

\bibitem[{Gururangan et~al.(2020)Gururangan, Marasovi{\'c}, Swayamdipta, Lo, Beltagy, Downey, and Smith}]{gururangan2020don}
Suchin Gururangan, Ana Marasovi{\'c}, Swabha Swayamdipta, Kyle Lo, Iz~Beltagy, Doug Downey, and Noah~A Smith. 2020.
\newblock \href {https://arxiv.org/abs/2004.10964} {Don't stop pretraining: Adapt language models to domains and tasks}.
\newblock \emph{arXiv preprint arXiv:2004.10964}.

\bibitem[{Hwang and Lee(2021)}]{hwang2021semi}
Hohyun Hwang and Younghoon Lee. 2021.
\newblock \href {https://aclanthology.org/2021.ranlp-1.67/} {Semi-supervised learning based on auto-generated lexicon using {XAI} in sentiment analysis}.
\newblock In \emph{Proceedings of the International Conference on Recent Advances in Natural Language Processing (RANLP 2021)}, pages 593--600.

\bibitem[{Jabbar et~al.(2019)Jabbar, Urooj, JunSheng, and Azeem}]{jabbar2019real}
Jahanzeb Jabbar, Iqra Urooj, Wu~JunSheng, and Naqash Azeem. 2019.
\newblock \href {https://ieeexplore.ieee.org/document/8743331} {Real-time sentiment analysis on e-commerce application}.
\newblock In \emph{2019 IEEE 16th international conference on networking, sensing and control (ICNSC)}, pages 391--396. IEEE.

\bibitem[{Joshi et~al.(2020)Joshi, Santy, Budhiraja, Bali, and Choudhury}]{joshi2020state}
Pratik Joshi, Sebastin Santy, Amar Budhiraja, Kalika Bali, and Monojit Choudhury. 2020.
\newblock \href {https://aclanthology.org/2020.acl-main.560/} {The state and fate of linguistic diversity and inclusion in the {NLP} world}.
\newblock \emph{arXiv preprint arXiv:2004.09095}.

\bibitem[{Kanana~Erastus and Kebeya(2018)}]{kanana2018functions}
Fridah Kanana~Erastus and Hilda Kebeya. 2018.
\newblock \href {https://link.springer.com/chapter/10.1007/978-3-319-64562-9_2} {Functions of urban and youth language in the new media: The case of {S}heng in {K}enya}.
\newblock \emph{African youth languages: New media, performing arts and sociolinguistic development}, pages 15--52.

\bibitem[{Learning(2006)}]{learning2006semi}
Semi-Supervised Learning. 2006.
\newblock \href {https://www.researchgate.net/publication/343972398_Semi-Supervised_Learning} {Semi-supervised learning}.
\newblock \emph{CSZ2006. html}, 5.

\bibitem[{Lee and Wang(2015)}]{lee2015emotion}
Sophia Lee and Zhongqing Wang. 2015.
\newblock \href {https://aclanthology.org/W15-3116/} {Emotion in code-switching texts: Corpus construction and analysis}.
\newblock In \emph{Proceedings of the Eighth SIGHAN workshop on chinese language processing}, pages 91--99.

\bibitem[{Lewis(2014)}]{lewis2014ethnologue}
M~Paul Lewis. 2014.
\newblock \href {https://www.sil.org/resources/archives/6133} {Ethnologue: Languages of the world}.
\newblock https://www.sil.org/about/endangered-languages/languages-of-the-world.

\bibitem[{Liu(2022)}]{liu2022sentiment}
Bing Liu. 2022.
\newblock \href {https://link.springer.com/book/10.1007/978-3-031-02145-9} {\emph{Sentiment Analysis and Opinion Mining}}.
\newblock Springer Nature.

\bibitem[{Liu et~al.(2019)Liu, Ott, Goyal, Du, Joshi, Chen, Levy, Lewis, Zettlemoyer, and Stoyanov}]{liu2019roberta}
Yinhan Liu, Myle Ott, Naman Goyal, Jingfei Du, Mandar Joshi, Danqi Chen, Omer Levy, Mike Lewis, Luke Zettlemoyer, and Veselin Stoyanov. 2019.
\newblock \href {https://arxiv.org/abs/1907.11692} {Roberta: A robustly optimized bert pretraining approach}.
\newblock \emph{arXiv preprint arXiv:1907.11692}.

\bibitem[{Matthew(2018)}]{matthew2018peters}
E~Matthew. 2018.
\newblock \href {https://aclanthology.org/N18-1202/} {Peters, mark neumann, mohit iyyer, matt gardner, christopher clark, kenton lee, luke zettlemoyer. deep contextualized word representations}.
\newblock In \emph{Proc. of NAACL}, volume~5.

\bibitem[{Mazrui(1995)}]{mazrui1995slang}
Alamin~M Mazrui. 1995.
\newblock \href {https://d-nb.info/1238150829/34} {Slang and code-switching: The case of {S}heng in {K}enya}.
\newblock \emph{Afrikanistische Arbeitspapiere: Schriftenreihe des K{\"o}lner Instituts f{\"u}r Afrikanistik}, (42):168--179.

\bibitem[{Medhat et~al.(2014)Medhat, Hassan, and Korashy}]{medhat2014sentiment}
Walaa Medhat, Ahmed Hassan, and Hoda Korashy. 2014.
\newblock \href {https://www.sciencedirect.com/science/article/pii/S2090447914000550} {Sentiment analysis algorithms and applications: A survey}.
\newblock \emph{Ain Shams engineering journal}, 5(4):1093--1113.

\bibitem[{Mohammad(2016)}]{mohammad2016practical}
Saif Mohammad. 2016.
\newblock \href {https://aclanthology.org/W16-0429/} {A practical guide to sentiment annotation: Challenges and solutions}.
\newblock In \emph{Proceedings of the 7th Workshop on Computational Approaches to Subjectivity, Sentiment and Social Media Analysis}, pages 174--179.

\bibitem[{Mohammad(2017)}]{mohammad2017word}
Saif~M Mohammad. 2017.
\newblock \href {https://arxiv.org/abs/1704.08798} {Word affect intensities}.
\newblock \emph{arXiv preprint arXiv:1704.08798}.

\bibitem[{Mohammad(2022)}]{mohammad2022ethics}
Saif~M Mohammad. 2022.
\newblock \href {https://direct.mit.edu/coli/article/48/2/239/109904/Ethics-Sheet-for-Automatic-Emotion-Recognition-and} {Ethics sheet for automatic emotion recognition and sentiment analysis}.
\newblock \emph{Computational Linguistics}, 48(2):239--278.

\bibitem[{Momanyi(2009)}]{momanyi2009effects}
Clara Momanyi. 2009.
\newblock \href {https://mail.jpanafrican.org/docs/vol2no8/2.8_EffectsOf.pdf} {The effects of'{S}heng'in the teaching of {K}iswahili in {K}enyan schools.}
\newblock \emph{Journal of Pan African Studies}.

\bibitem[{Muhammad et~al.(2023{\natexlab{a}})Muhammad, Abdulmumin, Ayele, Ousidhoum, Adelani, Yimam, Ahmad, Beloucif, Mohammad, Ruder et~al.}]{muhammad2023afrisenti}
Shamsuddeen~Hassan Muhammad, Idris Abdulmumin, Abinew~Ali Ayele, Nedjma Ousidhoum, David~Ifeoluwa Adelani, Seid~Muhie Yimam, Ibrahim~Sa'id Ahmad, Meriem Beloucif, Saif Mohammad, Sebastian Ruder, et~al. 2023{\natexlab{a}}.
\newblock \href {https://arxiv.org/abs/2302.08956} {Afrisenti: A twitter sentiment analysis benchmark for african languages}.
\newblock \emph{arXiv preprint arXiv:2302.08956}.

\bibitem[{Muhammad et~al.(2023{\natexlab{b}})Muhammad, Abdulmumin, Yimam, Adelani, Ahmad, Ousidhoum, Ayele, Mohammad, and Beloucif}]{muhammad2023semeval}
Shamsuddeen~Hassan Muhammad, Idris Abdulmumin, Seid~Muhie Yimam, David~Ifeoluwa Adelani, Ibrahim~Sa'id Ahmad, Nedjma Ousidhoum, Abinew Ayele, Saif~M Mohammad, and Meriem Beloucif. 2023{\natexlab{b}}.
\newblock \href {https://arxiv.org/abs/2304.06845} {Semeval-2023 task 12: sentiment analysis for african languages (afrisenti-semeval)}.
\newblock \emph{arXiv preprint arXiv:2304.06845}.

\bibitem[{Nagel(2018)}]{commoncrawlCommonCrawl}
Sebastian Nagel. 2018.
\newblock {C}ommon {C}rawl - {B}log - {I}ndex to {W}{A}{R}{C} {F}iles and {U}{R}{L}s in {C}olumnar {F}ormat --- commoncrawl.org.
\newblock \url{https://commoncrawl.org/blog/index-to-warc-files-and-urls}.
\newblock [Accessed 27-06-2024].

\bibitem[{Najafi et~al.(2019)Najafi, Maeda, Koyama, and Miyato}]{najafi2019robustness}
Amir Najafi, Shin-ichi Maeda, Masanori Koyama, and Takeru Miyato. 2019.
\newblock \href {https://proceedings.neurips.cc/paper_files/paper/2019/file/60ad83801910ec976590f69f638e0d6d-Paper.pdf} {Robustness to adversarial perturbations in learning from incomplete data}.
\newblock \emph{Advances in Neural Information Processing Systems}, 32.

\bibitem[{Naseem and Musial(2019)}]{naseem2019dice}
Usman Naseem and Katarzyna Musial. 2019.
\newblock \href {https://ieeexplore.ieee.org/document/8978072} {Dice: Deep intelligent contextual embedding for twitter sentiment analysis}.
\newblock In \emph{2019 International Conference on Document Analysis and Recognition (ICDAR)}, pages 953--958. IEEE.

\bibitem[{Nasukawa and Yi(2003)}]{nasukawa2003sentiment}
Tetsuya Nasukawa and Jeonghee Yi. 2003.
\newblock \href {https://dl.acm.org/doi/10.1145/945645.945658} {Sentiment analysis: Capturing favorability using natural language processing}.
\newblock In \emph{Proceedings of the 2nd International Conference on Knowledge Capture}, pages 70--77.

\bibitem[{Nielsen(2011)}]{nielsen2011new}
Finn~{\AA}rup Nielsen. 2011.
\newblock \href {https://arxiv.org/abs/1103.2903} {A new anew: Evaluation of a word list for sentiment analysis in microblogs}.
\newblock \emph{arXiv preprint arXiv:1103.2903}.

\bibitem[{Niesler and De~Wet(2008)}]{niesler2008accent}
Thomas Niesler and Febe De~Wet. 2008.
\newblock \href {https://www.isca-archive.org/odyssey_2008/niesler08_odyssey.pdf} {Accent identification in the presence of code-mixing.}
\newblock In \emph{Odyssey}, page~27.

\bibitem[{Niesler et~al.(2018)}]{niesler2018first}
Thomas Niesler et~al. 2018.
\newblock \href {https://aclanthology.org/L18-1451.pdf} {A first south african corpus of multilingual code-switched soap opera speech}.
\newblock In \emph{Proceedings of the Eleventh International Conference on Language Resources and Evaluation (LREC 2018)}.

\bibitem[{Ogueji et~al.(2021)Ogueji, Zhu, and Lin}]{ogueji2021small}
Kelechi Ogueji, Yuxin Zhu, and Jimmy Lin. 2021.
\newblock \href {https://aclanthology.org/2021.mrl-1.11/} {Small data? no problem! exploring the viability of pretrained multilingual language models for low-resourced languages}.
\newblock In \emph{Proceedings of the 1st Workshop on Multilingual Representation Learning}, pages 116--126.

\bibitem[{Olatunji et~al.(2023)Olatunji, Afonja, Yadavalli, Emezue, Singh, Dossou, Osuchukwu, Osei, Tonja, Etori et~al.}]{olatunji2023afrispeech}
Tobi Olatunji, Tejumade Afonja, Aditya Yadavalli, Chris~Chinenye Emezue, Sahib Singh, Bonaventure~FP Dossou, Joanne Osuchukwu, Salomey Osei, Atnafu~Lambebo Tonja, Naome Etori, et~al. 2023.
\newblock \href {https://direct.mit.edu/tacl/article/doi/10.1162/tacl_a_00627/118796} {Afrispeech-200: Pan-african accented speech dataset for clinical and general domain asr}.
\newblock \emph{Transactions of the Association for Computational Linguistics}, 11:1669--1685.

\bibitem[{Ortigosa et~al.(2014)Ortigosa, Mart{\'\i}n, and Carro}]{ortigosa2014sentiment}
Alvaro Ortigosa, Jos{\'e}~M Mart{\'\i}n, and Rosa~M Carro. 2014.
\newblock \href {https://www.sciencedirect.com/science/article/abs/pii/S0747563213001751} {Sentiment analysis in facebook and its application to e-learning}.
\newblock \emph{Computers in human behavior}, 31:527--541.

\bibitem[{Otundo and Grice(2022)}]{otundo2022intonation}
Billian~Khalayi Otundo and Martine Grice. 2022.
\newblock \href {https://drive.google.com/file/d/1Uk7c8qzZj8aZvlLfihw41MYAOLGSecYD/view} {Intonation in advice-giving in kenyan english and kiswahili}.
\newblock \emph{Proceedings of Speech Prosody 2022}, pages 150--154.

\bibitem[{Pang et~al.(2002)Pang, Lee, and Vaithyanathan}]{pang2002thumbs}
Bo~Pang, Lillian Lee, and Shivakumar Vaithyanathan. 2002.
\newblock \href {https://arxiv.org/abs/cs/0205070} {Thumbs up? sentiment classification using machine learning techniques}.
\newblock \emph{arXiv preprint cs/0205070}.

\bibitem[{Pang et~al.(2008)Pang, Lee et~al.}]{pang2008opinion}
Bo~Pang, Lillian Lee, et~al. 2008.
\newblock \href {https://www.cs.cornell.edu/home/llee/omsa/omsa.pdf} {Opinion mining and sentiment analysis}.
\newblock \emph{Foundations and Trends{\textregistered} in information retrieval}, 2(1--2):1--135.

\bibitem[{Pham et~al.(2023)Pham, Pham, Nguyen, Nguyen, and Dinh}]{pham2023semi}
Viet~H Pham, Thang~M Pham, Giang Nguyen, Long Nguyen, and Dien Dinh. 2023.
\newblock \href {https://arxiv.org/abs/2304.00557} {Semi-supervised neural machine translation with consistency regularization for low-resource languages}.
\newblock \emph{arXiv preprint arXiv:2304.00557}.

\bibitem[{Piergallini et~al.(2016)Piergallini, Shirvani, Gautam, and Chouikha}]{piergallini2016word}
Mario Piergallini, Rouzbeh Shirvani, Gauri~Shankar Gautam, and Mohamed Chouikha. 2016.
\newblock \href {https://aclanthology.org/W16-5803/} {Word-level language identification and predicting codeswitching points in swahili-english language data}.
\newblock In \emph{Proceedings of the second workshop on computational approaches to code switching}, pages 21--29.

\bibitem[{Poplack(2000)}]{poplack2000toward}
Shana Poplack. 2000.
\newblock \href {https://eric.ed.gov/?id=ED214394} {Toward a typology of code-switching}.
\newblock \emph{L. WEI ({\'e}d.), The bilingualism reader. London, New York: Routeledge}, pages 221--255.

\bibitem[{Raffel et~al.(2020)Raffel, Shazeer, Roberts, Lee, Narang, Matena, Zhou, Li, and Liu}]{raffel2020exploring}
Colin Raffel, Noam Shazeer, Adam Roberts, Katherine Lee, Sharan Narang, Michael Matena, Yanqi Zhou, Wei Li, and Peter~J Liu. 2020.
\newblock \href {https://www.jmlr.org/papers/v21/20-074.html} {Exploring the limits of transfer learning with a unified text-to-text transformer}.
\newblock \emph{Journal of Machine Learning Research}, 21(140):1--67.

\bibitem[{Ren and Quan(2012)}]{ren2012linguistic}
Fuji Ren and Changqin Quan. 2012.
\newblock \href {https://link.springer.com/article/10.1007/s10799-012-0138-5} {Linguistic-based emotion analysis and recognition for measuring consumer satisfaction: an application of affective computing}.
\newblock \emph{Information Technology and Management}, 13:321--332.

\bibitem[{Saeki et~al.(2023)Saeki, Zen, Chen, Morioka, Wang, Zhang, Bapna, Rosenberg, and Ramabhadran}]{saeki2023virtuoso}
Takaaki Saeki, Heiga Zen, Zhehuai Chen, Nobuyuki Morioka, Gary Wang, Yu~Zhang, Ankur Bapna, Andrew Rosenberg, and Bhuvana Ramabhadran. 2023.
\newblock \href {https://ieeexplore.ieee.org/document/10095702?denied=} {Virtuoso: Massive multilingual speech-text joint semi-supervised learning for text-to-speech}.
\newblock In \emph{ICASSP 2023-2023 IEEE International Conference on Acoustics, Speech and Signal Processing (ICASSP)}, pages 1--5. IEEE.

\bibitem[{Sanh et~al.(2019)Sanh, Debut, Chaumond, and Wolf}]{sanh2019distilbert}
Victor Sanh, Lysandre Debut, Julien Chaumond, and Thomas Wolf. 2019.
\newblock \href {https://arxiv.org/abs/1910.01108} {Distilbert, a distilled version of bert: smaller, faster, cheaper and lighter}.
\newblock \emph{arXiv preprint arXiv:1910.01108}.

\bibitem[{Santy et~al.(2021)Santy, Srinivasan, and Choudhury}]{santy2021bertologicomix}
Sebastin Santy, Anirudh Srinivasan, and Monojit Choudhury. 2021.
\newblock \href {https://aclanthology.org/2021.adaptnlp-1.12/} {{BERTologiCoMix}: How does code-mixing interact with multilingual {BERT}?}
\newblock In \emph{Proceedings of the Second Workshop on Domain Adaptation for NLP}, pages 111--121.

\bibitem[{Scotton(1993)}]{scotton1993social}
Carol~Myers Scotton. 1993.
\newblock \href {https://academic.oup.com/book/48387} {\emph{Social motivations for codeswitching: Evidence from Africa}}.
\newblock Clarendon Press.

\bibitem[{Singh and Singh(2022)}]{singh2022low}
Salam~Michael Singh and Thoudam~Doren Singh. 2022.
\newblock \href {https://www.sciencedirect.com/science/article/abs/pii/S0957417422013513} {Low resource machine translation of {E}nglish--{M}anipuri: A semi-supervised approach}.
\newblock \emph{Expert Systems with Applications}, 209:118187.

\bibitem[{Strassel and Tracey(2016)}]{strassel2016lorelei}
Stephanie Strassel and Jennifer Tracey. 2016.
\newblock \href {https://aclanthology.org/L16-1521/} {Lorelei language packs: Data, tools, and resources for technology development in low resource languages}.
\newblock In \emph{Proceedings of the Tenth International Conference on Language Resources and Evaluation (LREC'16)}, pages 3273--3280.

\bibitem[{Suttles and Ide(2013)}]{suttles2013distant}
Jared Suttles and Nancy Ide. 2013.
\newblock \href {https://link.springer.com/chapter/10.1007/978-3-642-37256-8_11} {Distant supervision for emotion classification with discrete binary values}.
\newblock In \emph{International Conference on Intelligent Text Processing and Computational Linguistics}, pages 121--136. Springer.

\bibitem[{Taboada et~al.(2011)Taboada, Brooke, Tofiloski, Voll, and Stede}]{taboada2011lexicon}
Maite Taboada, Julian Brooke, Milan Tofiloski, Kimberly Voll, and Manfred Stede. 2011.
\newblock \href {https://aclanthology.org/J11-2001/} {Lexicon-based methods for sentiment analysis}.
\newblock \emph{Computational linguistics}, 37(2):267--307.

\bibitem[{Terblanche et~al.(2024)Terblanche, Olaleye, and Marivate}]{terblanche2024prompting}
Michelle Terblanche, Kayode Olaleye, and Vukosi Marivate. 2024.
\newblock \href {https://aclanthology.org/2024.sigul-1.33.pdf} {Prompting towards alleviating code-switched data scarcity in under-resourced languages with gpt as a pivot}.
\newblock \emph{arXiv preprint arXiv:2404.17216}.

\bibitem[{Thara and Poornachandran(2018)}]{thara2018code}
S~Thara and Prabaharan Poornachandran. 2018.
\newblock \href {https://ieeexplore.ieee.org/document/8554413} {Code-mixing: A brief survey}.
\newblock In \emph{2018 International Conference on Advances in Computing, Communications and Informatics (ICACCI)}, pages 2382--2388. IEEE.

\bibitem[{Thomas et~al.(2013)Thomas, Seltzer, Church, and Hermansky}]{thomas2013deep}
Samuel Thomas, Michael~L Seltzer, Kenneth Church, and Hynek Hermansky. 2013.
\newblock \href {https://ieeexplore.ieee.org/document/6638959} {Deep neural network features and semi-supervised training for low resource speech recognition}.
\newblock In \emph{2013 IEEE International Conference on Acoustics, Speech and Signal Processing}, pages 6704--6708. IEEE.

\bibitem[{Vaswani et~al.(2017)Vaswani, Shazeer, Parmar, Uszkoreit, Jones, Gomez, Kaiser, and Polosukhin}]{vaswani2017attention}
Ashish Vaswani, Noam Shazeer, Niki Parmar, Jakob Uszkoreit, Llion Jones, Aidan~N Gomez, {\L}ukasz Kaiser, and Illia Polosukhin. 2017.
\newblock \href {https://proceedings.neurips.cc/paper_files/paper/2017/file/3f5ee243547dee91fbd053c1c4a845aa-Paper.pdf} {Attention is all you need}.
\newblock \emph{Advances in neural information processing systems}, 30.

\bibitem[{Vo and Collier(2013)}]{vo2013twitter}
Bao-Khanh~Ho Vo and NIGEL Collier. 2013.
\newblock \href {http://www.ijcla.org/2013-1/IJCLA-2013-1-pp-159-173-09-Twitter.pdf} {Twitter emotion analysis in earthquake situations.}
\newblock \emph{Int. J. Comput. Linguistics Appl.}, 4(1):159--173.

\bibitem[{Vo and Zhang(2015)}]{vo2015target}
Duy-Tin Vo and Yue Zhang. 2015.
\newblock \href {https://www.ijcai.org/Proceedings/15/Papers/194.pdf} {Target-dependent twitter sentiment classification with rich automatic features}.
\newblock In \emph{Twenty-fourth International Joint Conference on Artificial Intelligence}.

\bibitem[{Wang et~al.(2019)Wang, Pruksachatkun, Nangia, Singh, Michael, Hill, Levy, and Bowman}]{wang2019superglue}
Alex Wang, Yada Pruksachatkun, Nikita Nangia, Amanpreet Singh, Julian Michael, Felix Hill, Omer Levy, and Samuel Bowman. 2019.
\newblock \href {https://arxiv.org/abs/1905.00537} {Superglue: A stickier benchmark for general-purpose language understanding systems}.
\newblock \emph{Advances in Neural Information Processing systems}, 32.

\bibitem[{Wang et~al.(2024)Wang, Adelani, Agrawal, Masiak, Rei, Briakou, Carpuat, He, Bourhim, Bukula et~al.}]{wang2024afrimte}
Jiayi Wang, David Adelani, Sweta Agrawal, Marek Masiak, Ricardo Rei, Eleftheria Briakou, Marine Carpuat, Xuanli He, Sofia Bourhim, Andiswa Bukula, et~al. 2024.
\newblock \href {https://arxiv.org/abs/2311.09828} {Afrimte and africomet: Enhancing comet to embrace under-resourced african languages}.
\newblock In \emph{Proceedings of the 2024 Conference of the North American Chapter of the Association for Computational Linguistics: Human Language Technologies (Volume 1: Long Papers)}, pages 5997--6023.

\bibitem[{Winata et~al.(2022)Winata, Aji, Yong, and Solorio}]{winata2022decades}
Genta~Indra Winata, Alham~Fikri Aji, Zheng-Xin Yong, and Thamar Solorio. 2022.
\newblock \href {https://aclanthology.org/2023.findings-acl.185/} {The decades progress on code-switching research in nlp: A systematic survey on trends and challenges}.
\newblock \emph{arXiv preprint arXiv:2212.09660}.

\bibitem[{Zamani et~al.(2016)Zamani, Abas, and Amin}]{zamani2016eye}
H~Zamani, A~Abas, and MKM Amin. 2016.
\newblock \href {https://jtec.utem.edu.my/jtec/article/view/1415} {Eye tracking application on emotion analysis for marketing strategy}.
\newblock \emph{Journal of Telecommunication, Electronic and Computer Engineering (JTEC)}, 8(11):87--91.

\bibitem[{Zhu et~al.(2015)Zhu, Kiros, Zemel, Salakhutdinov, Urtasun, Torralba, and Fidler}]{zhu2015aligning}
Yukun Zhu, Ryan Kiros, Rich Zemel, Ruslan Salakhutdinov, Raquel Urtasun, Antonio Torralba, and Sanja Fidler. 2015.
\newblock \href {https://arxiv.org/abs/1506.06724} {Aligning books and movies: Towards story-like visual explanations by watching movies and reading books}.
\newblock In \emph{Proceedings of the IEEE international conference on computer vision}, pages 19--27.

\end{thebibliography}

\clearpage

\appendix

\section{Appendix}

\subsection{Language Detection}

\begin{table}[ht]
\begin{tabular}{lll}
\toprule
\parbox{.75in}{\textbf{Language Code}} & \textbf{Occurrences} & \textbf{Language} \\
\midrule
en & 29845 & English \\
id & 3288 & Indonesian \\
sw & 624 & Swahili \\
no & 192 & Norwegian \\
da & 119 & Danish \\
tr & 95 & Turkish \\
nl & 81 & Dutch \\
af & 73 & Afrikaans \\
de & 71 & German \\
ca & 55 & Catalan \\
so & 46 & Somali \\
sv & 34 & Swedish \\
et & 26 & Estonian \\
tl & 15 & Tagalog (Filipino) \\
hu & 14 & Hungarian \\
fr & 14 & French \\
es & 10 & Spanish \\
hr & 9 & Croatian \\
it & 8 & Italian \\
cy & 8 & Welsh \\
fi & 6 & Finnish \\
pl & 4 & Polish \\
sl & 3 & Slovenian \\
lt & 3 & Lithuanian \\
ro & 3 & Romanian \\
\bottomrule
\end{tabular}
\caption{Count of language detection in the RideKE dataset}
\label{tab:my_label}
\end{table}


\subsection{Tweets Per Location}

\begin{figure}[ht]
  \includegraphics[width=\columnwidth]{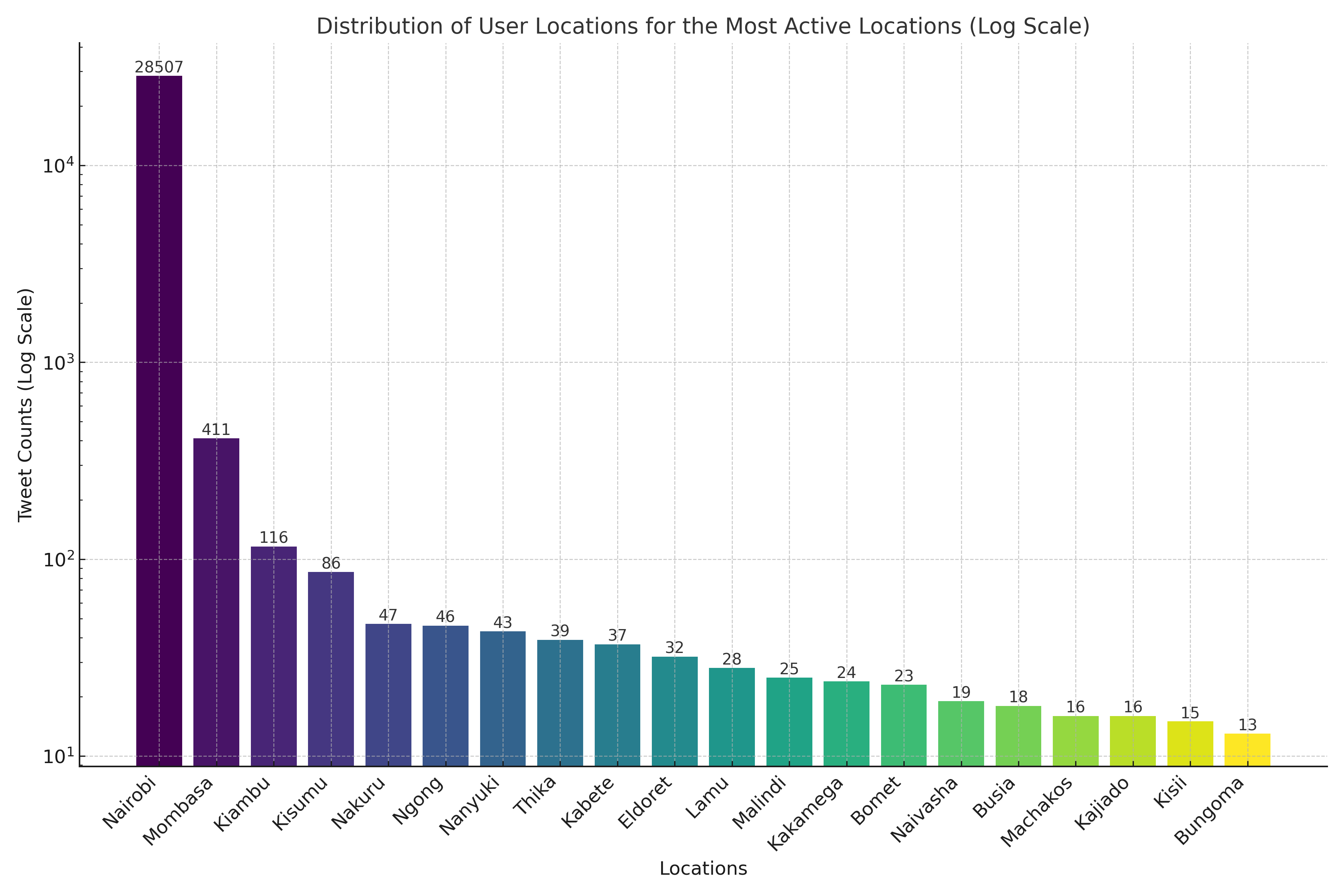}
  \caption{Number of tweets per location on a logarithmic scale. Nairobi appears to be the most active location per dataset.}
  \label{fig:number of tweets per location using logarithmic scale}
\end{figure}

\subsection{\rule{0pt}{6ex}Sheng-to-English Sample Sentences}

\begin{table}[ht]
\centering
\begin{tabular}{ll}
\toprule
\textbf{Sheng} & \textbf{English Translation} \\
\midrule
dere anadai  &  Driver demands \\
kuna some people eating  & people benefitting \\
ferry slay queens & Ferry divas \\
Mmemulikwaa oya  & on the spotlight ! \\
Mhesh  & honorable sir  \\
wazungu’s& white people \\
sikwembe ya Yesu & strong faith in Jesus \\
Hiyo pesa ni kadonye & That’s little money \\
fare noma & Expensive fare \\
kuweka ngata & To fuel\\
\bottomrule
\end{tabular}
\caption{Sheng to English Example Sentences}
\label{table:sheng_english}
\end{table}

\onecolumn
\newpage

\subsection{Annotation Guidelines}
\label{sec:appendix_annotation}
\begin{table*}[!htbp]
\centering
\small
\begin{tabular}{|l|p{\dimexpr\textwidth-2\tabcolsep-2in\relax}|} 
\hline
\textbf{Aspect} & \textbf{Details} \\
\hline
Title & Annotation Guidelines for RHS Conversation on Twitter \\
\hline
Task & Annotating emotions in tweets related to RHS experiences \\
\hline
Annotation Process & 
\begin{itemize} \setlength{\itemsep}{1pt} \setlength{\parskip}{0pt} \setlength{\parsep}{0pt}
    \item Emotion Definition: Annotators accurately identify and label the predominant emotion expressed in each tweet based on the emotional tone conveyed by the text.
    \item Keyword Identification: Pay attention to keywords or phrases that suggest the presence of a particular emotion.
    \item Context Matters: Consider the tweet's context, including any relevant hashtags, mentions, or user profiles, for a better understanding of the emotional context.
    \item Tweet Length: Emotions can be expressed differently in short and long tweets.
\end{itemize} \\
\hline
Emotion Labels Guidelines & 
\begin{enumerate}\setlength{\partopsep}{0pt} \setlength{\itemsep}{0pt} \setlength{\parskip}{0pt}
    \item \textbf{Anger:} Label when the tweet expresses frustration, annoyance, resentment, or strong displeasure toward RHS, drivers, or related issues. Look for keywords and tone indicative of anger. Keywords: angry, furious, annoyed, upset. Example: "Terrible experience with Uber driver! He was rude and refused to follow the GPS directions \#Angry".
    \item \textbf{Happy:} Label when the tweet reflects joy, satisfaction, contentment, or delight regarding RHS experiences. Look for expressions of happiness, appreciation, or positive feedback. Keywords: happy, delighted, thrilled, satisfied. Example: ``Just had the best ride ever with the friendliest driver! \#HappyCustomer \#GreatService''
    \item \textbf{Fear:} Label when the tweet expresses anxiety, worry, concern, or fear about RHS safety, incidents, or perceived risks. Identify cues of fear or apprehension. Keywords: afraid, scared, worried, nervous. Example: "My ride is taking an unfamiliar route, and I'm getting worried. Is this safe? \#Fear"
    \item \textbf{Suprise:} Label when the tweet indicates astonishment, amazement, or unexpected reactions to RHS experiences.Keywords: surprised, shocked, amazed, unexpected. Example: "Wow, my driver gave me a free upgrade to a luxury car! \#Surprised 
    \item \textbf{Love:} Label when the tweet reflects affection, appreciation, or strong positive emotions toward RHS, drivers, or related aspects. Look for expressions of love or admiration. Keywords: love, adore, appreciate, grateful. Example: "Wow, my driver gave me a free upgrade to a luxury car! \#Surprised \#Love"
    \item \textbf{Frustration:} Label when the tweet expresses dissatisfaction, irritation, or being fed up with RHS issues. Identify cues of frustration and annoyance. Keyword: frustrated, annoyed, fed up, irritated. Example: "Been waiting for my ride for ages. This is so frustrating!\#Frustrated \#LateAgain"
   
     \item \textbf{Neutral:} Label when the tweet does not exhibit any strong emotional sentiment or when the emotion is unclear or ambiguous. Use this label sparingly and only when other emotions are not evident. Example: "Just booked my ride for tomorrow morning. \#RideHail \#PlanningAhead
   
  \end{enumerate} \\
\hline
Quality Control & Monitor inter-annotator agreement to ensure consistency among annotators. Resolve disagreements through discussion and clarification. \\
\hline
Privacy and Ethical Considerations & Respect user privacy and report any offensive content appropriately. \\
\hline
\end{tabular}
\caption{Annotation guidelines for ride-hailing service conversation emotions on Twitter}
\label{tab:emotion_guidelines}
\end{table*}

\newpage

\subsection{Sample dataset structure}

\newcommand{\0}[1]{{\bf \begin{tabular}{@{}l@{}}#1\end{tabular}}}

\begin{table}[!htbp]
\centering
\resizebox{\textwidth}{!}{%
\begin{tabular}{|l|l|p{5cm}|l|l|l|l|l|l|l|l|l|}
\hline
\textbf{Keyword} & \textbf{Date} & \textbf{Tweets} & \0{reply\\count} & \0{retweet\\count} & \0{like\\count} & \textbf{verified} & \0{user\\followers} & \0{user\\following} & \0{user\\tweets} & \0{user\\location} & \textbf{country} \\ \hline
\#UBER-Kenya & 2023-04-10 & Did Nairobi ask you to double Nairobi fare price ? That's how Uber Kenya and bolt steal from us here. & 1 & 0 & 0 & 0 & 2104 & 981 & 23173 & Mombasa & Kenya \\ \hline
\#UBER-Kenya & 2023-03-30 & Uber Kenya made an order that was cancelled by a restaurant but I've already paid. How do I follow up on my refund? & 1 & 0 & 0 & 0 & 946 & 975 & 4642 & Nairobi & Kenya \\ \hline
\#UBER-Kenya & 2023-03-30 & Uber Kenya made an order that was cancelled by a restaurant but I've already paid. How do I follow up on my refund? & 1 & 0 & 0 & 0 & 946 & 975 & 4642 & Nairobi & Kenya \\ \hline
\#UBER-Kenya & 2023-04-02 & Uber is losing the Kenyan market to Nairobi apps, customers are tired of being asked by drivers where in Nairobi they are going. Nairobi apps show Nairobi drivers where the customer is, where is going and price hence drivers will decide to accept or decline the request. & 2 & 0 & 0 & 0 & 46 & 297 & 817 & Nairobi & Kenya \\ \hline
\#UBER-Kenya & 2023-03-27 & Uber Kenya how can your driver click not paid when he was paid? And Nairobi is proof of payment? & 2 & 0 & 0 & 0 & 5744 & 1338 & 208846 & Nairobi & Kenya \\ \hline
\#BOLT-Kenya & 2023-04-06 & Hello, thanks for writing in. Kindly do reach out to us via kenyabolt.eu and a member of our team will respond and assist accordingly. & 0 & 0 & 0 & 1 & 15093 & 447 & 16966 & Nairobi & Kenya \\ \hline
\#BOLT-Kenya & 2023-01-16 & Let's have an honest conversation here...this morning you lowered the base category to 8ksh per kilometer. We all know that fuel is still very high. What method did you use to reach this point, Did you involve drivers about the & 1 & 0 & 3 & 0 & 14 & 87 & 82 & Nairobi & Kenya \\ \hline
\#BOLT-Kenya & 2022-11-19 & If you don't communicate. Let us as drivers do what we feel like doing. Because bolt Kenya is manner less. & 0 & 0 & 0 & 0 & 11 & 69 & 66 & Nairobi & Kenya \\ \hline
\#BOLT-Kenya & 2019-10-27 & How come Bolt Kenya does not have an active customer service line for queries? & 0 & 0 & 0 & 0 & 23 & 180 & 35 & Nairobi & Kenya \\ \hline
\#BOLT-Kenya & 2019-05-20 & Thanks to Boltkenya been arriving at my studio sessions and interviews on time and with comfort. You too can enjoy this service by simply downloading boltkenya and using my code FEMIONEBOLT to get kshs250 off & 0 & 0 & 3 & 0 & 40629 & 528 & 24042 & Nairobi & Kenya \\ \hline
\#LITTLECAB & 2022-07-31 & Now \#Littlecab will not allow me to cancel a ride I did not take until I pay. Exhausting! & 0 & 0 & 0 & 0 & 636 & 2222 & 4085 & Nairobi & Kenya \\ \hline
\#LITTLECAB & 2022-07-31 & And they let me have their driver. The security officer at \#Carnivorekenya says that they do not verify the drivers. What is the whole point of telling us to use \#littlecab if you have no relationship with them. Just destroyed my whole experience attending a beautiful musical. & 1 & 0 & 0 & 0 & 636 & 2222 & 4085 & Nairobi & Kenya \\ \hline
\#LITTLECAB & 2022-05-10 & Use \#Littlecab. These other Apps are foreign and exploitive. & 0 & 0 & 1 & 0 & 480 & 987 & 4740 & Mombasa & Kenya \\ \hline
\#LITTLECAB & 2020-12-23 & Why do we always encounter cabs from \#LittleCab that arrive with different number plates from what is registered in your system? While I don't board them in principle for security concerns, it may one day be costly for a desperate client & 2 & 0 & 0 & 0 & 4712 & 3044 & 20683 & Nairobi & Kenya \\ \hline
\end{tabular}%
}
\caption{Original sample of the tweets data structure}
\label{tab:tweet_data_analysis}
\end{table}

\end{document}